\documentclass[twoside,11pt]{article}

\usepackage{jmlr2e}
\usepackage[utf8]{inputenc} 
\usepackage{subfigure}

\usepackage{algpseudocode}
\usepackage{algorithm}
\usepackage{algorithmicx}

\usepackage{isma}
\definecolor{mygreen}{RGB}{76,153,0}
\usepackage{booktabs} 

\usepackage{sidecap}
\usepackage{caption}
\newcommand{\opt}{^*}

\newcommand{\st}{\mathcal{S}}

\usepackage{lastpage}
\ShortHeadings{title}{Lemhadri, Ruan, Abraham, Tibshirani}
\ShortHeadings{LassoNet}{A Neural Network with Feature Sparsity}
\firstpageno{1}

\begin{document}

\title{LassoNet: A Neural Network with Feature Sparsity}

\author{\name Ismael Lemhadri \email lemhadri@stanford.edu \\
      \addr Department of Statistics, Stanford University, Stanford, U.S.A.
      \AND
      \name Feng Ruan \email fengruan@berkeley.edu \\
      \addr Department of Statistics,
      University of California,
      Berkeley, USA
      \AND
      \name Louis Abraham \email louis.abraham@yahoo.fr \\
      \addr Gematria Technologies, London, U.K.
      \AND
      \name Robert Tibshirani \email tibs@stanford.edu \\
      \addr Departments of Biomedical Data Sciences, and Statistics, Stanford University, Stanford, U.S.A.}

\editor{}

\maketitle

\begin{abstract}
Much work has been done recently to make neural networks more interpretable, and one approach is to arrange for the network to use only a subset of the available features. In linear models, Lasso (or $\ell_1$-regularized) regression assigns zero weights to the most irrelevant or redundant features, and is widely used in data science. However the Lasso only applies to linear models. Here we introduce LassoNet, a neural network framework with global feature selection. Our approach achieves feature sparsity by adding a  skip (residual) layer and  allowing a feature to participate in any  hidden layer only if its skip-layer representative is active. Unlike other approaches to feature selection for neural nets, our method uses a modified objective function with constraints, and so integrates feature selection with the parameter learning directly. As a result, it delivers an entire regularization path of solutions with a range of feature sparsity. We apply LassoNet to a number of real-data problems and find that it significantly outperforms state-of-the-art methods for feature selection and regression. LassoNet uses projected proximal gradient descent, and generalizes directly to deep networks. It can be implemented by adding just a few lines of code to a standard neural network.
\end{abstract}


\section{Introduction}
\subsection{Background}
In many problems of interest, much of the information in the features is irrelevant for predicting the responses and only a small subset is informative.
Feature selection methods provide insight into the relationship between features and an outcome while simultaneously reducing the computational expense of downstream learning by removing features that are redundant or noisy.

With high-dimensional data sets becoming ever more prevalent, feature selection has seen widespread usage across a variety of real-world tasks, including disease detection from protein data, speech data and object recognition \citep{wulfkuhle2003proteomic,cai2018feature,li2017feature}.
The benefits of feature selection include reducing experimental costs, enhancing interpretability, computational speed up, memory reduction and even improving model generalization on unseen data \citep{min2014feature,ribeiro2016should,chandrashekar2014survey}.
For example, feature selection is especially valuable in biomedical studies where the data with the full set of features is expensive or difficult to collect, as it can alleviate the need to measure irrelevant or redundant features, and allows to identify a small set of features while maintaining prediction performance $-$ this can significantly save on future data collection costs.
While feature selection methods have been extensively studied in the setting of linear regression (e.g. LASSO), identifying relevant features for neural networks remains an open challenge.

As a motivating example, consider a data set that consists of the expression levels of various proteins across tissue samples.
Such measurements are increasingly carried out to assist with disease diagnosis, as biologists measure a large number of proteins with the aim of discriminating between disease classes.
Yet, it remains expensive to conduct all of the measurements that are needed to fully characterize proteomic diseases.
It is natural to ask:  \textit{Are there redundant or unnecessary features?  What are the most effective and representative features to characterize the disease?}
Furthermore, when a small number of proteins are selected, their biological relationship with the target diseases is more easily identified. 
These "marker" proteins thus provide additional scientific understanding of the problem.

\begin{figure}
    \centering
    \includegraphics[width=0.6\linewidth]{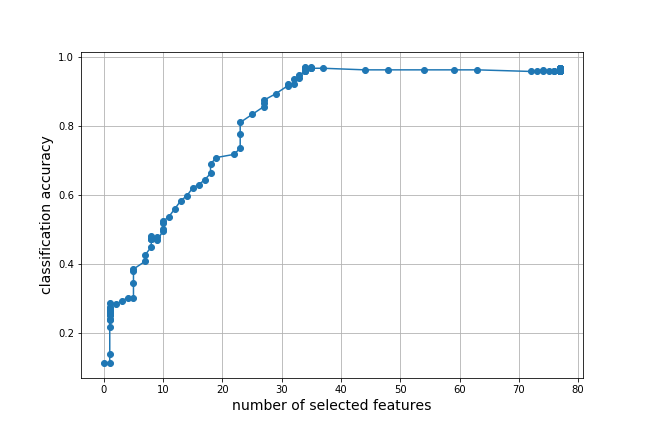}
    \caption{
    \textbf{Feature selection path produced by our method on the MICE Protein Dataset} \citep{higuera2015self}.
    Considering the cost of proteomic measurements, a trade-off between the number of features kept and statistical performance is often desirable.
    In this example, the method captures 70\% of the signal with about 20\% of the features. This allows to narrow down the list of important features, making the conclusions of the prediction task more actionable. More details on the MICE Protein Dataset are available in Section \ref{sec:experiments}.
    }
    \label{fig:mice-path}
\end{figure}

Figure \ref{fig:mice-path} shows an example of feature selection path produced by our method on the MICE Protein Dataset \citep{higuera2015self}, which contains protein expression levels of normal and trisomic mice exposed to different experimental conditions. The curve is steeply concave and typical, which explains why feature selection is often a key pre-processing step in many machine learning tasks. We see that only about 35 proteins are needed to obtain maximal classification accuracy.

The outline of this paper is as follows. First, we discuss related work on feature selection.  Section \ref{sec:problem-formulation} formulates the problem from a non-parametric model selection perspective. Section \ref{sec:proposal} introduces our main proposal, and the optimization is presented in Section \ref{sec:optimization}. In Section \ref{sec:experiments}, we conduct an experiments with several real-world datasets. Sections \ref{sec:unsupervised} and \ref{sec:matrix-completion} offer extensions of LassoNet to the unsupervised learning setting and to the matrix completion problem, respectively.
Finally, Section \ref{sec:other-extensions} discusses other possible extensions.

\subsection{Related Works}
Feature selection methods can generally be divided into three groups: filter, wrapper and embedded methods. 
\begin{itemize}
    \item Filter methods operate independently of the choice of the predictor by selecting individual features that maximize the desired criteria. For example, the popular Fisher score \citep{ gu2012generalized} selects features such that in the data space spanned by the selected features, the distances between data points in different classes are as large as possible, while the distances between data points in the same class are as small as possible.
    Relatedly, \cite{song2012feature} propose to maximize Maximum Mean Discrepancy, \textit{i.e.} the difference between between-class distance and within-class distance.
    Filter methods select features independently of the learning method to be used, and this is a major limitation.
For example, since filter methods evaluate individual features, they generally do not detect features that participate mainly in interactions with other features.
    \item Wrapper methods use learning algorithms to evaluate subsets of features based on their predictive power. 
    For example, the recently proposed HSIC-LASSO \citep{yamada2014high} uses kernel learning to discover non-linear feature interactions.
    In another line of work, \cite{chen2017kernel} use the well-known kernel dimension reduction algorithm to perform feature selection.
    
    \item Similarly to wrapper methods, embedded methods use specific predictors to select features, and are generally able to detect interactions and redundancies among features. However, embedded methods tend to do so more efficiently as they combine feature selection and learning into a single problem.
    A well-known example is the Lasso \citep{Ti96}, which can be used to select features for regression by varying the strength of $l_1$ regularization.
    The limitation of lasso, however, is that it only applies to linear models.
    Recently, \cite{feng2017sparse} proposed an input-sparse neural network, where the input weights are penalized using the group Lasso penalty. As will become evident in Section \ref{sec:proposal}, our proposed method extends and generalizes this approach in a natural way.
\end{itemize}

The literature on feature selection is vast and encompasses many fields.
We do not provide a comprehensive survey here, but focus on popular methods in machine learning instead.
We point the reader to \cite{Guyon03} for a more in-depth review of the feature selection literature.

We also note that in certain problems of interest, feature selection starts by looking for the appropriate \textit{representation basis}. 
As a consequence, one might transform the data into this basis before applying feature selection algorithms. 
For example, it is commonplace in signal processing problems to convert time series data to the frequency domain first.
In this work, we consider this step to be pre-processing.
We will therefore assume that the correct basis is already available.
However, in Section \ref{sec:other-extensions} we discuss the more general problem of simultaneously learning and selecting features.

\subsection{Proposed Method}
\label{sec:proposed-method}
We propose a new approach that extends Lasso regression and its feature sparsity to feed-forward neural networks.
We call our procedure \textit{LassoNet}.
The method is designed so that only a subset of the features are used by the network.
Our procedure uses an input-to-output skip-layer (residual) connection that allows a feature to have non-zero weights in a hidden unit only if its skip-layer connection is active.

\begin{figure}
\begin{minipage}{.5\textwidth}
  \centering
  \includegraphics[width=1\linewidth]{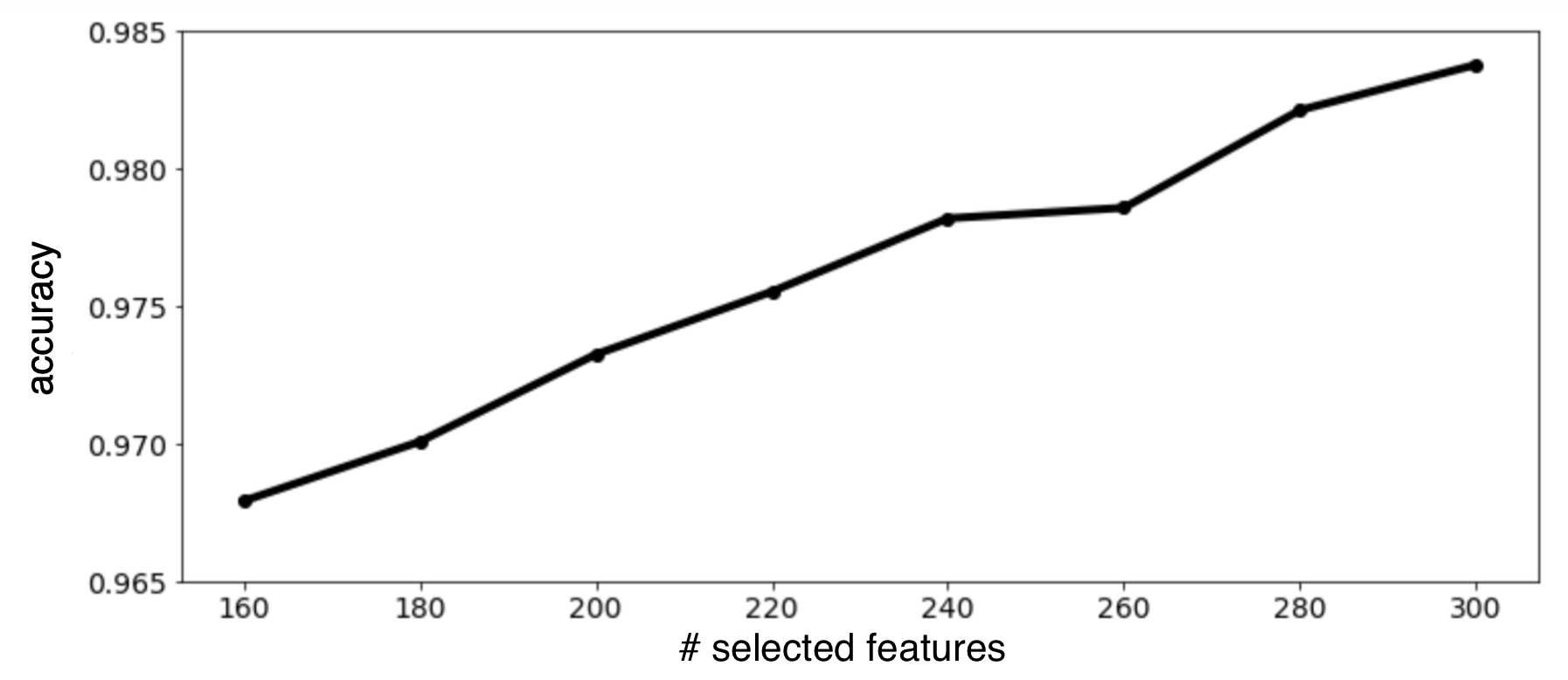}
\end{minipage}\hfill
\begin{minipage}{.25\textwidth}
  \centering
  \includegraphics[width=1\linewidth]{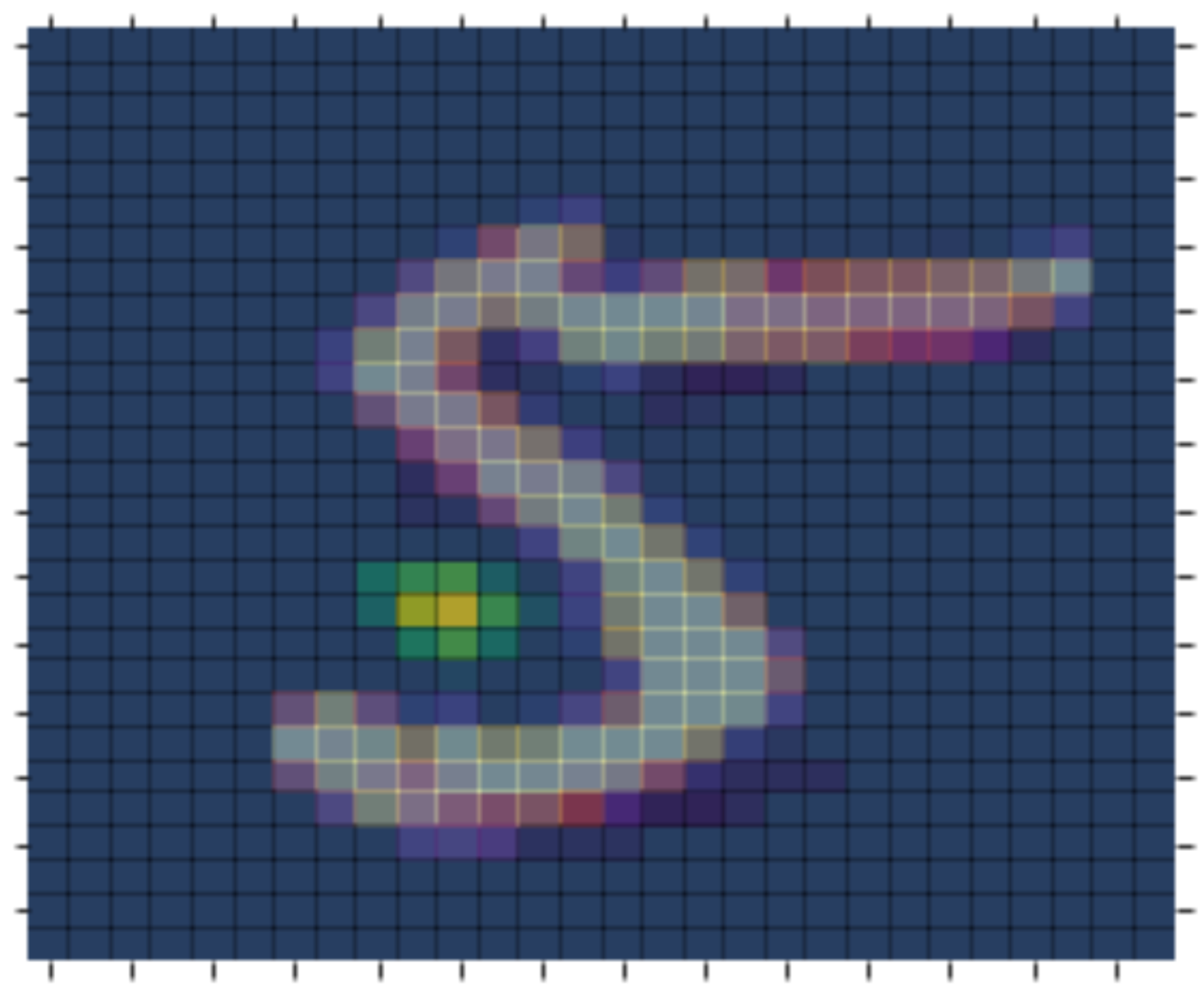}
\end{minipage}%
\begin{minipage}{.25\textwidth}
  \centering
  \includegraphics[width=1\linewidth]{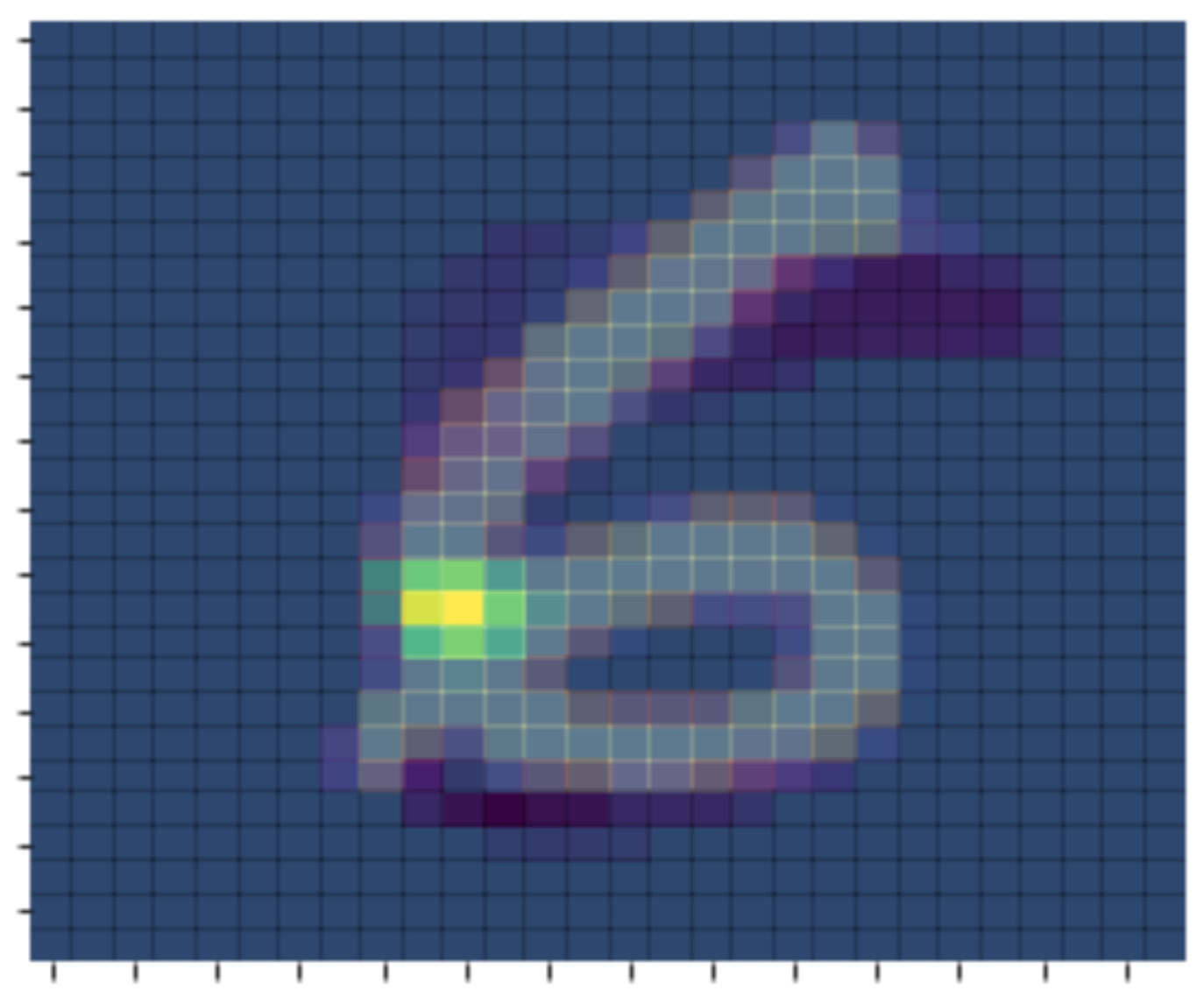}
\end{minipage}%
\caption{
\textbf{Demonstrating LassoNet on the MNIST dataset.} Here, we show the results of using LassoNet to simultaneously select informative pixels and classify digits 5 and 6 from the MNIST dataset.
\textbf{Leftmost graph:} The classification accuracy by number of selected
features
\textbf{Right 2 images:} Individual pixel importance for the model with 220 active features.
Here, pixel importance refers to the mean increase in digit 6's predicted probability when setting it to maximum intensity. 
Lighter colors indicate higher importance, with yellow highest and dark blue lowest. Superimposed are two sample digits.
This confirms that the bottom left pixels of digit 6 are the most important.
}
\label{fig:mnist}
\end{figure}

The linear and nonlinear components are optimized jointly, allowing to capture arbitrary nonlinearity. As we show through experiments in Section \ref{sec:experiments}, this leads to lower classification errors on real-world datasets compared to the aforementioned methods.
A visual example of results from our method is shown in Fig. \ref{fig:mnist}, where LassoNet selects the most informative pixels on a subset of the MNIST dataset, and classifies the original images with high accuracy.
We test LassoNet on a variety of datasets, and find that it generally outperforms state-of-the-art methods
for feature selection and regression.

We have made the code for our algorithm and experiments available on a public website \footnote{\href{Code available at https://lassonet.ml}{https://lassonet.ml}}.

\section{Problem Formulation}
\label{sec:problem-formulation}
We now describe the problem of global feature selection. Although global feature selection is relevant for both supervised and unsupervised settings, we describe here the supervised case, which is the focus of this paper, and defer discussion of the unsupervised case to Section \ref{sec:unsupervised}.

We assume a data-generating model $p(\textbf{x},y)$ over a $d$-dimensional space, where $\textbf{x} \in \mathbb{R}^d$ is the covariate and $y$ is the response, such as class labels.
The goal is to find the best function $f\opt(\textbf{x})$ for predicting $y$.
We emphasize that the problem of learning $f^*$ is non-parametric, so that for example no linear or quadratic restriction is assumed.
We seek to minimize the empirical reconstruction error:
\begin{equation}
\label{eqn:risk-minimization}
    \min_{f\in \mathcal{F}, S}~~\widehat{\E}[\ell(f(\textbf{x}_S),y) ]
\end{equation}
where $S \subseteq \{1,2\ldots d\} $ is a subset of features and $\ell$ is a loss function specified by the user. For example, in a univariate regression problem, the function class might be the set of all linear functions, and $\ell$ is a loss function such as $\ell(f(\textbf{x}),y)=(y-f(x))^2$.
The principal difficulty in solving~\eqref{eqn:risk-minimization} is due to the combinatorial nature of the minimization---the choice of possible subsets $S$ grows exponentially in $d$, making the problem NP-hard even for simple choices of $f$, such as linear regression \citep{amaldi1998approximability}, and exhaustive search is intractable if the number of features is large. In addition, the function class $\mathcal{F}$ needs to exhibit strong expressive power---that is, we seek to develop a method that can approximate the solution for any given class of functions, from linear regression to deep fully-connected neural networks.

\section{Our proposal: LassoNet}
\label{sec:proposal}

\subsection{Background and notation}
\label{sec:notation}
Here we choose $\mathcal{F}$ to be the class of residual feed-forward neural networks:
\begin{equation*}
    \mathcal{F} = \left\{f \equiv f_{\theta,W}: x \mapsto
    \theta ^T \textbf{x} + g_W(\textbf{x}) \right\},
\end{equation*}
where the width and depth of the network are arbitrary.
Residual networks are known to be easier to train \citep{he2016deep}. Furthermore, they act as universal approximators to many function classes \citep{raghu2017expressive, lin2018resnet}.

For the reader's convenience, we collect key notation and background here. Throughout the paper: 
\begin{itemize}
    \item $n$ denotes the total number of training observations;
    \item $d$ denotes the data dimension;
    \item $g_W$ denotes a feed-forward network with weights $W$ (fully connected in our examples);
    \item $K$ denotes the number of units in the first hidden layer;
    \item $W^{(1)} \in \mathbb{R}^{d\times K}$ denotes the weights in the first hidden layer, and $\theta \in \mathbb{R}^d$ denotes the weights in the residual layer;
    \item $L(\theta,W) = \frac{1}{n}\sum_{i=1}^n \ell (f_{\theta,W}(\textbf{x}_i), y_i)$ is the empirical loss\footnote{We have removed the dependence of $L$ on $(\textbf{x}_i,y_i)$ for notational convenience} on the training data set, consisting of samples $\{\textbf{x}_i, y_i \}_1^n$) and $\ell$ is the loss function such as squared-error, cross-entropy (deviance) or misclassification rate.
    \item $\st_{\lambda}(x) = \sign(x) \cdot \max\left\{|x| - \lambda, 0\right\}$ is the soft-thresholding operator.
\end{itemize}

\begin{SCfigure}
    \includegraphics[width=0.45\linewidth]{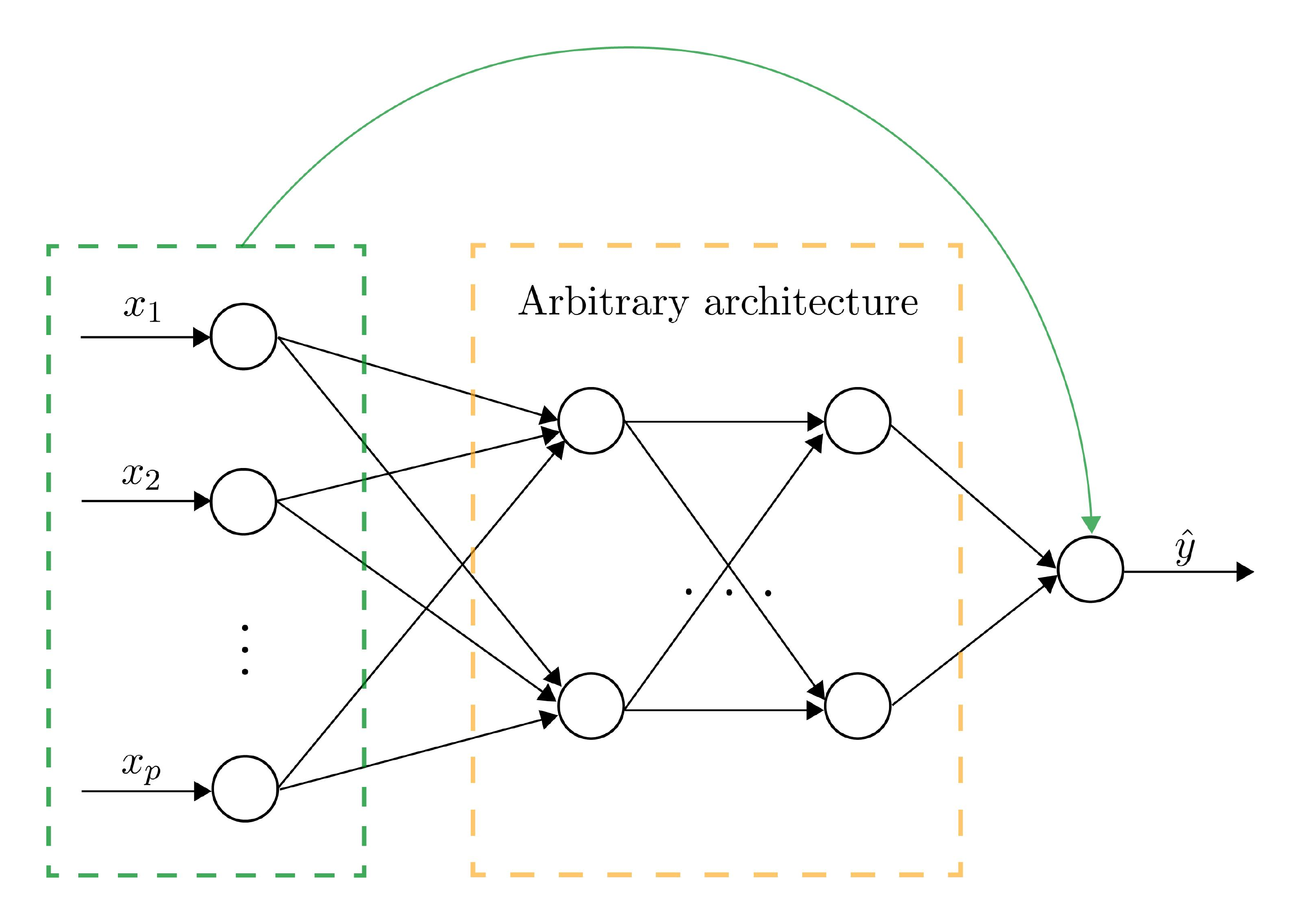}
    \caption{\textbf{LassoNet architecture} \\
The architecture of LassoNet consists of a single residual connection, shown in \textcolor{mygreen}{green} and an arbitrary feed-forward neural network, shown in \textbf{black}. The residual layer and the first hidden layer jointly pass through a hierarchical soft-thresholding optimizer.}
\label{fig:architecture}
\end{SCfigure}

The general architecture of LassoNet is illustrated in Fig. \ref{fig:architecture}. The method consists of two main ingredients:
\begin{enumerate}
    \item A \textbf{penalty} is introduced to the original empirical risk minimization that encourages feature sparsity.
    The formulation transforms the combinatorial search to a continuous search by varying the level of the penalty.
    
    \item A \textbf{proximal algorithm} is applied in a mathematically elegant way, so that it admits a simple and efficient implementation on top of back-propagation. 
    The method can be implemented by adding just a few lines of code to a standard neural network. The mathematical derivation of this algorithm is detailed in Section 5. 
\end{enumerate}

\subsection{Formulation}
The LassoNet objective function is defined as

    \begin{equation}
    \label{eqn:lassonet}
    \begin{split}
    &\minimize_{\theta,W}
        \;
        L(\theta,W) + \lambda \norm{\theta}_1
    	 \\
    &\subjectto ~\normbig{W^{(1)}_j}_\infty \le  M |\theta_j|,\: j = 1,\ldots,d.
    \end{split}
    \end{equation}
where the loss function $L(\theta, W)$ was defined above in Section \ref{sec:notation}, and $W_j^{(1)}$ denotes the weights for feature $j$ in the first hidden layer. 
We emphasize that our goal is not just to sparsify the network, but to do so in a structured way that selects the relevant input features for the entire network. 
Since the network is feed-forward, we do not need to penalize the weights in remaining hidden layers.

The key idea in our approach is the constraint
\begin{eqnarray*}
|W^{(1)}_{jk}| \leq M\cdot |\theta_j| , \: k = 1,\ldots, K
\end{eqnarray*}
This budgets the total amount of non-linearity involving feature $j$ according to the relative effect importance of $X_j$ as a main effect.
An immediate consequence is that $W_j = 0$ if $\theta_{j} = 0$. In other words, feature $j$ {\em does not participate in the overall network if the skip-layer weight  $\theta_{j}$ is zero}. Hence control of the sparsity of the skip layer weights gives complete control of the feature sparsity in the network.

In the extreme case where $M = 0$, all the hidden units are inactive and only the skip connection remains.
That is, the formulation recovers exactly the Lasso.
In the other extreme (by letting $M \to +\infty$), one recovers a standard unregularized feed-forward neural network.

This formulation has several benefits.
First, it promotes the linear component of the signal above the nonlinear one and uses it to guide feature sparsity. 
Such a strategy is not new, and bears close resemblance to the \textit{hierarchy principle} which has been extensively studied in statistics \citep{choi2010variable, radchenko2010variable,LH2014,she2016, yan2017}.
In addition, the formulation leverages the expressive power of residual neural networks \citep{he2016deep}.
Finally, by tying every feature to a single coefficient (the linear component), our formulation provides a natural framework for feature selection.

One added benefit of the formulation is that the linear and non-linear components are fitted \textit{simultaneously}, allowing the network to capture arbitrary nonlinearity in the data. If the best fitting model would have $\normbig{W^{(1)}_{j}}_\infty$ large but $|\theta_j|$ only moderate, this can be accommodated with a reasonable choice of $M$. 
Furthermore, Fig. \ref{fig:toy} suggests that the demand for hierarchy is analogous to the demand for sparsity—--a form of “regularization.”

Training LassoNet involves two operations. 
First, a vanilla gradient descent step is applied to all model parameters. Then, a hierarchical proximal operator is applied to the input layer pair $(\theta,W^{(1)})$.
This sequential operation makes the procedure extremely simple to implement in popular machine learning frameworks, and requires only modifying a few lines of code from a standard residual network.
The procedure is summarized in Alg. \ref{alg:lassonet}.

An added benefit of the method is its computational efficiency. 
The LassoNet regularization path can be trained at a cost that is essentially that of training a \textit{single} model. 
This is achieved with the use of warm starts in a specific direction, as outlined in Section \ref{sec:warm-starts}.

\begin{algorithm}
    \caption{Training LassoNet}
    \label{alg:lassonet}
	\begin{algorithmic}[1]
	\State \textbf{Input:} training dataset $X \in \mathbb{R}^{n \times d}$, training labels $Y$, feed-forward neural network $g_W(\cdot)$, number of epochs $B$, hierarchy multiplier $M$, path multiplier $\epsilon$, learning rate $\alpha$
	\State Initialize and train the feed-forward network on the loss $L(\theta,W)$
	\State Initialize the penalty, $\lambda=\lambda_0$, and the number of active features, $k=d$
	\While {$k > 0$}
	    \State Update $\lambda \leftarrow (1+\epsilon)\lambda$
	    \For{$b \in \{1\ldots B\}$}
    	    \State Compute gradient of the loss w.r.t to $(\theta,W)$ using back-propagation
    	    \State Update $\theta \leftarrow \theta - \alpha \nabla_{\theta}L$ and $W \leftarrow W - \alpha \nabla_{_W}L$
    	    \State Update $(\theta, W^{(1)}) \leftarrow 
    	    \textsc{Hier-Prox}(
    	    \theta, W^{(1)}, \alpha \lambda, M)$
	    \EndFor
    	\State Update $k$ to be the number of non-zero coordinates of $\theta$
	\EndWhile
	\State
	where \textsc{Hier-Prox} is defined in Alg. \ref{alg:hier-prox}
	\end{algorithmic} 
\end{algorithm}

\subsection{Hyper-parameter tuning}
LassoNet has two hyper-parameters: 
\begin{itemize}
    \item the $\ell_1$-penalty coefficient, $\lambda$, controls the complexity of the fitted model; higher values of $\lambda$ encourage sparser models;
    \item the hierarchy coefficient, $M$, controls the relative strength of the linear and nonlinear components.
    
\end{itemize}

It may be difficult to set the hierarchy coefficient without expert knowledge on the domain or task. We can circumvent this problem by treating the hierarchy coefficient as a hyper-parameter. We may use a naive search, which exhaustively evaluates the accuracy for the predefined hyper-parameter candidates with a validation dataset. This procedure can be performed in parallel.

\section{Optimization}
\label{sec:optimization}

\subsection{Warm starts: a path from dense to sparse}
\label{sec:warm-starts}
The technique of warm starts is very effective in optimizing models over an entire regularization path. For example, 
this technique is employed in Lasso $\ell_1$-regularized linear regression \citep{FHT2010}.
In this approach, optimization is carried out for each fixed value of $\lambda$ on a logarithmic scale from sparse to dense, and using the solution from the previous $\lambda$ as a warm start for the next. This is effective, since the sparse models are easier to optimize and the sparse solution is often of main interest.

\begin{figure}
\centering
\begin{subfigure}{}
  \centering
  \includegraphics[width=.52\linewidth, height=2.35in]{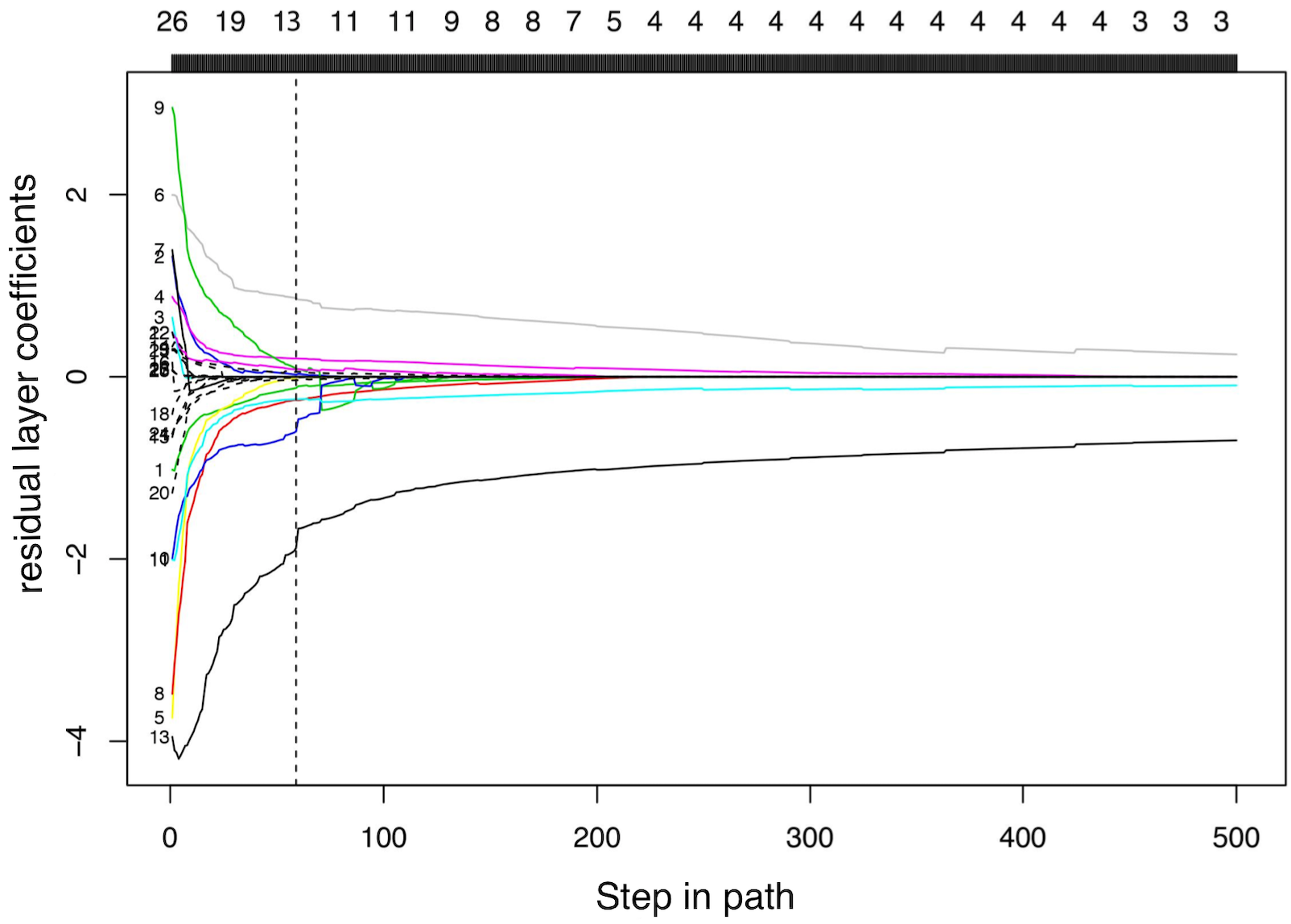}
\end{subfigure}%
\begin{subfigure}{}
  \centering
  \includegraphics[width=.42\linewidth, height=2.3in]{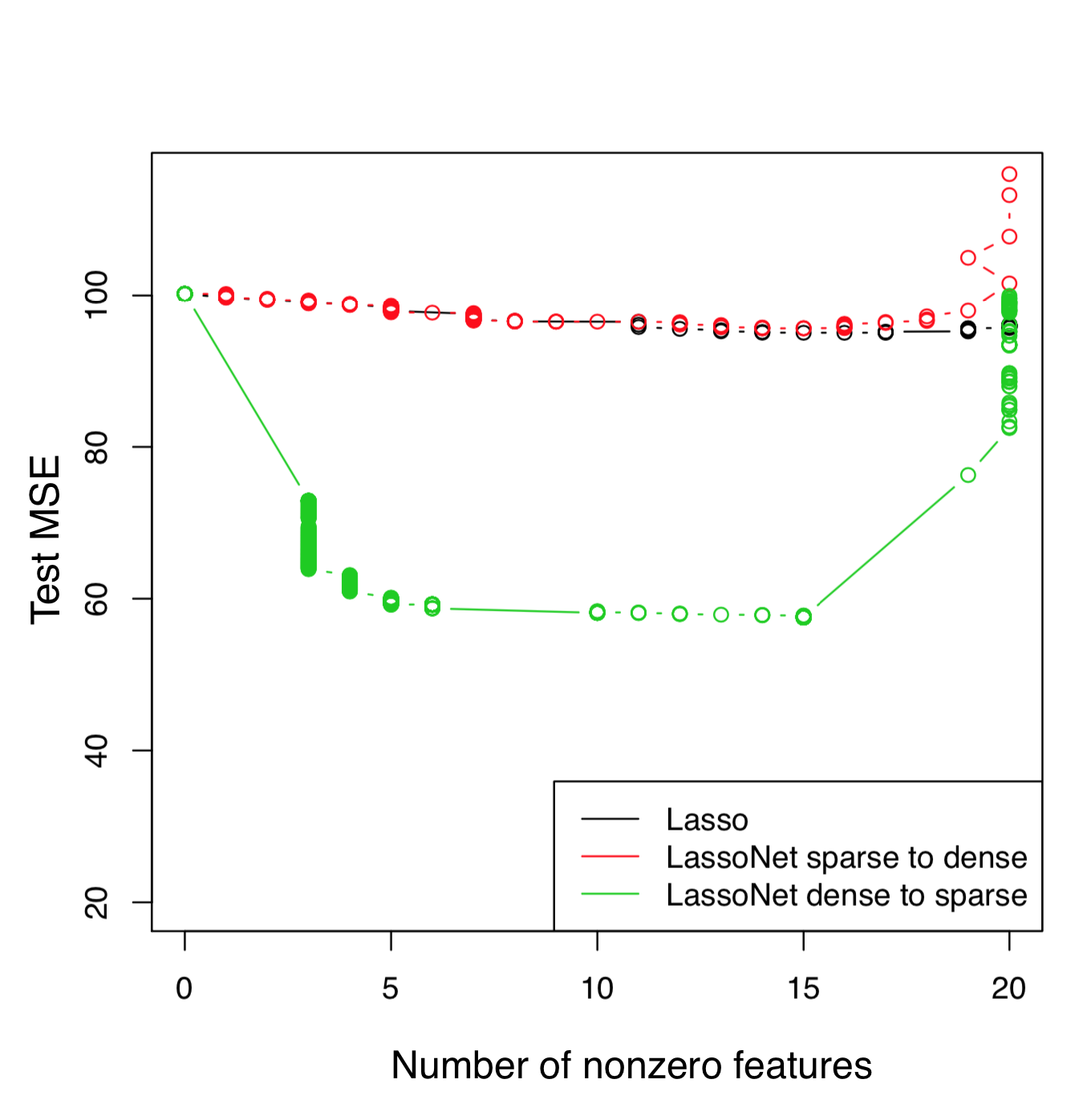}
\end{subfigure}
\caption{
\textbf{Left:} The path of residual coefficients for the Boston housing dataset.
We augmented the Boston Housing dataset from $p=13$ features with $13$ additional Gaussian noise features (corresponding to the broken lines).
The number of features selected by LassoNet is indicated along the top.
LassoNet achieves the minimum test error (at the vertical broken line) at 13 predictors. Upon inspection of the resulting model, 12 of the 13 selected features correspond to the true predictors, confirming the model's ability to perform controlled feature selection.
\textbf{Right:} Comparing two kinds of initialization.
The test errors for Lasso and LassoNet using both the sparse-to-dense and dense-to-sparse strategies are shown. The dense-to-sparse strategy achieves superior performance, confirming the importance of a dense initialization in order to efficiently explore the optimization landscape.
}
\label{fig:toy}
\end{figure}

Somewhat surprisingly, to optimize LassoNet we find that a \emph{dense-to-sparse} warm start approach is far more effective than a sparse-to-dense approach, in the sense that the former approach returns models that generalize better than those returned from the latter.
This phenomenon is illustrated in Fig. \ref{fig:toy}, where the standard sparse-to-dense approach gets caught in local minima with poor generalization ability.
On the other hand, the dense-to-sparse approach leverages the favorable generalization properties of the dense solution and preserves them after drifting into sparse local minima.

\subsection{Hierarchical proximal optimization}
\label{subsec:hier-prox}

The objective is optimized using proximal gradient descent, as outlined in Alg. 1. The key novelty is a numerically efficient algorithm for the proximal inner loop.  We call the proposed algorithm \textsc{Hier-Prox} and detail it in Alg. 2.  Underlying its development is the derivation of equivalent optimality conditions that completely characterize the \emph{global} solution of the \emph{non-convex} minimization problem defining the proximal operator. As it turns out, the inner loop is decomposable across features. As we show in Appendix \ref{proof}, \textsc{Hier-Prox} finds the global minimum of an optimization problem of the form

\begin{equation*}
\begin{split}
&\minimize_{b \in \R, W \in \R^K}
	\half (v-b)^2 + \half \ltwo{u-W}^2 + \lambda |b|, \\
&\subjectto ~\norm{W}_\infty \le  M |b|
\end{split}
\end{equation*}

Remarkably, the complexity of \textsc{Hier-Prox} is controlled by $O(p \cdot \log p)$, where $p$ is the total number of parameters being updated (\textit{i.e.} $p = dK + d$)
This overhead is negligible compared to the computation of the gradients with respect to the same parameters. 
Furthermore, implementing the optimizer is straightforward in most standard deep learning frameworks.
We provide more information about our implementation in Appendix \ref{appendix:experiments}.

\begin{algorithm}
    \caption{Hierarchical Proximal Operator}
    \label{alg:hier-prox}
	\begin{algorithmic}[1]
	\Procedure{Hier-Prox}{$\theta,W^{(1)};\lambda,M$}
	\For{$j \in \{1,\ldots, d\}$}
	    \State Sort the entries of $W^{(1)}_{j}$ into $|W^{(1)}_{(j,1)}| \geq \ldots \geq |W^{(1)}_{(j,K)}|$
	    \For{$m \in \{0, \ldots, K\}$}
	    
	        \State Compute 
	        $w_m := \frac{M}{1+mM^2} \cdot
    		\st_{\lambda} \Big(|\theta_j| + M\cdot \sum_{i=1}^m 
    	|W^{(1)}_{(j,i)}|\Big)$
    	\EndFor
    \State Find $\tilde{m}$, the first $m \in \{0, \ldots, K\}$ such that $|W^{(1)}_{(j,m+1)}|\leq w_m \leq |W^{(1)}_{(j,m)}|$
	\State $\Tilde{\theta}_j \gets \frac{1}{M}\cdot \sign (\theta_j) \cdot w_{\tilde{m}}$
	\State $\Tilde{W}^{(1)}_j \gets \sign (W^{(1)}_j)\cdot \min(w_{\tilde{m}},|W^{(1)}_j|)$
	\EndFor
	\State
	\Return $(\Tilde{\theta},\Tilde{W}^{(1)})$
	\EndProcedure
	\State Notation: $d$ denotes the number of features; $K$ denotes the size of the first hidden layer.
	\State Conventions:
	Ln. 6, $W^{(1)}_{(j,K+1)} = 0$, $W^{(1)}_{(j,0)} =+\infty$;
	Ln. 9, minimum and absolute value are applied coordinate-wise.
    \end{algorithmic}
\end{algorithm}

\subsection{Computational Complexity}
\label{subsec:complexity}
In most existing hierarchical models, computation remains a major challenge. 
Indeed, the complex nature of the regularizers used to enforce hierarchy prevents most current optimization algorithms from scaling with $d$.
In contrast, training LassoNet is performed at an attractive computational cost. Namely:
\begin{itemize}
    \item The bulk of the computational cost occurs when training the dense network;
    \item Subsequently, training over the $\lambda$ path is computationally cheap. By leveraging warm starts and the efficient \textsc{Hier-Prox} solver, the method effectively prunes the dense model. In practice, predictions across consecutive solutions in the path are usually close, which explains the speed-ups we observe in our experiments.
\end{itemize}

The use of warm starts dramatically reduces the number of rounds of gradient descent needed during each iteration, as the solution with penalty $\lambda$ is often very similar to the solution with penalty $(1+\epsilon)\lambda$.
This added benefit distinguishes LassoNet from many competing feature selection methods, which require advance knowledge of the optimal number of features to select, and do not exhibit any computational savings over the path of features.
Finally, the computational complexity of the method improves with hardware acceleration and parallelization techniques commonplace in deep learning.

\subsection{Bias due to regularization}
The use of the $\ell_1$ penalty slightly biases the weights of the model downwards. 
This is not exclusive to LassoNet, but is a well-known property of all $\ell_1$ regularized models such as Lasso \citep{lee2016exact}.
If the goal is to perform feature selection, this has no consequence on the training of LassoNet.
If the goal is to get optimal predictive performance, it can be helpful to de-bias LassoNet.
Our approach to do so is by re-training the network from scratch, restricting the input features to the subset of interest and zeroing out all others. 
In practice, this re-fitting tends to improve test-time performance. 
In addition, its computational overhead is relatively minor: (1) it is only needed after model selection, and therefore need not be performed on the entire regularization path;  (2) the training time of the reduced model is relatively light if one uses the previous (biased) model as initialization in the gradient descent.

\section{Experiments}
\label{sec:experiments}

In this section, we show experimental results on real-world datasets.

\subsection{Data Sets}
\label{subsec:data}
These datasets are drawn from several domains including protein data, image data and voice data, and have all been used for benchmarking feature selection methods in prior literature \citep{abid2019concrete}\footnote{The dataset descriptions were provided by these authors.} (the size of the datasets can be found in Table \ref{table:experiments}):

\begin{itemize}

    \item \textbf{Mice Protein Dataset} consists of protein expression levels measured in the cortex of normal and trisomic mice who had been exposed to different experimental conditions. Each
    feature is the expression level of one protein. \footnote{This data set has a few missing values, which we impute by column-means.}

    \item \textbf{MNIST and MNIST-Fashion} consist of 28-by-28 grayscale images of hand-written digits and clothing items, respectively. We choose these datasets because they are widely known in the machine learning community. Although these are image datasets, the objects in each image are centered, which means we can meaningfully treat each 784 pixels in the image as a separate feature.
    
    \item \textbf{ISOLET} consists of preprocessed speech data of people speaking the names of the letters in the English alphabet, and is widely used as a benchmark in the feature selection literature. Each feature is one of the 617 quantities produced as a result of preprocessing, including spectral coefficients and sonorant features.
    
    \item \textbf{COIL-20} consists of centered grayscale images of 20 objects. Images of the objects were taken at pose intervals of 5 degrees amounting to 72 images for each object. During preprocessing, the images were resized to produce 20-by-20 images, with each feature being one of the 400 pixels.
    
    \item \textbf{Smartphone Dataset for Human Activity Recognition} consists of sensor data collected from a smartphone mounted on subjects while they performed several activities such as walking upstairs, standing and laying. Each feature represents one of the 561 raw or processed quantities from the sensors on the phone.

\end{itemize}

\subsection{Methodology}
 We compare LassoNet with several supervised feature selection methods mentioned in the Related Works Section, including HSIC-LASSO and the Fisher Score. 
 We also include principal feature analysis (PFA), a popular method for selecting discrete features based on PCA, proposed by \cite{lu2007feature}.
 Where available, we made use of the \verb|scikit-feature| implementation \citep{li2018feature} of each method.
 Fig. \ref{fig:isolet} shows the results on the ISOLET data set , which is widely used as a benchmark in prior feature selection literature \citep{xue2020multiple, doquet2019agnostic}.
 
 We benchmarked each feature selection method with varying number of features. Although LassoNet is an integrated procedure --- simultaneously performing feature selection and learning, most other methods are not, and therefore we explore the use of the selected feature set as input into two separate downstream learners.
 
 In our experiments, we also include, as an upper-bound on performance, reconstruction methods that are not restricted to choosing individual features. In experiments with decoders, we use a standard feed-forward auto-encoder with all the input features, and in tree-based learners, we use equivalent full trees.

For every task, we run each algorithm being evaluated to extract the $k$ features selected.
We measure classification accuracy by passing the resulting matrix $X_S$ to a one-hidden-layer feed-forward network and to an extremely randomized trees classifier \citep{geurts2006extremely}, a variant of random forests that has been used with feature selection methods in prior literature \citep{drotar2015experimental}.
We chose these two classifiers to emphasize the inherent value of the selected features: since decoders and random forests are very different classifiers by nature, Figure \ref{fig:isolet} (for the ISOLET data set) indicates that the selected features are intrinsically informative (regardless of the classifier).

For all of the experiments, we use Adam optimizer with a learning rate of $10^{-3}$ to train the initial dense model. Then, we use vanilla gradient descent with momentum equal to $0.9$ on the regularization path. See Appendix D for the architecture of the networks.
For LassoNet, we did not use the network that was learned during training, but re-trained the feed-forward network from scratch. We divide each data set randomly into train, validation and test with a 70-10-20 split. 
The number of neurons in the hidden layer of the feed-forward network was varied within $[k/3,2k/3,k,4k/3]$, and the network with the highest validation accuracy was selected and measured on the test set. 

\subsection{Results}

The resulting classification errors are shown in Table \ref{table:experiments} for the decoder network, and in Appendix \ref{table:experiments_tree} for the tree-based classifier.
Overall, we find that our method is the strongest performer in the large majority of cases.
While occasionally more than one method achieves the best accuracy, we find that our method either ties or outperforms the remaining methods in all instances, suggesting that our hierarchical objective may be widely applicable for different learning tasks.

\begin{figure}
    \centering
    \includegraphics[width=1.\linewidth]{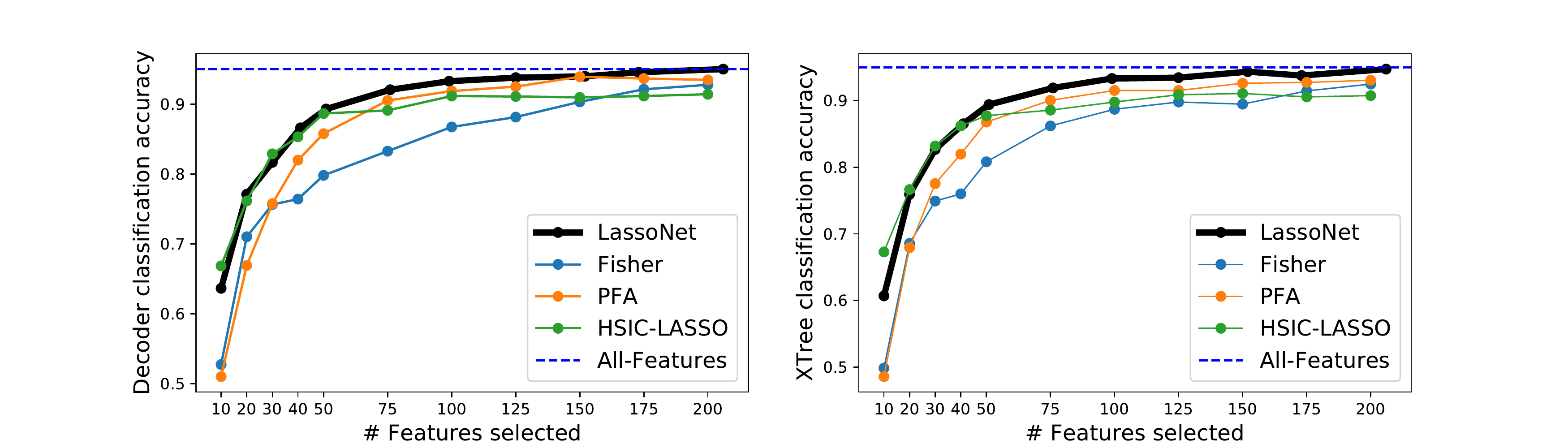}
    \caption{\textbf{Results on the ISOLET dataset.}
Here, we compare LassoNet to other feature selection methods using a 1-hidden layer neural network (\textit{left}) and an Extremely Randomized Trees (a variant of random forests) classifier (\textit{right}).
We find that across all values of $k$ tested, and for both learners, LassoNet has highest classification accuracy.}
\label{fig:isolet}
\end{figure}

In addition to Table \ref{table:experiments}, Figure \ref{fig:isolet} shows the classification accuracies for varying number of features on the ISOLET data set.
We also report the corresponding figures for the other data sets in Appendix \ref{app:experiments}, with similar results and conclusions.

\begin{table}
  \centering
  \begin{tabular}{cccccccc}
 \midrule
    Dataset & $(n,d)$ & \# Classes & \textit{All-Features} & Fisher & HSIC-Lasso & PFA & LassoNet\\
    \midrule
    Mice Protein & (1080, 77) & 8 & \textit{0.990} & 0.944 & \textbf{0.958} & 0.939 &  \textbf{0.958} \\
    MNIST & (10000, 784) & 10 & \textit{0.928} & 0.813 & 0.870 & \textbf{0.873}&  \textbf{0.873}	 \\	
    MNIST-Fashion & (10000, 784) & 10 & \textit{0.833} & 0.671 & 0.785 & 0.793 &  \textbf{0.800} \\	
     ISOLET & (7797, 617) & 26 & \textit{0.953} & 0.793 & 0.877 & 0.863&  \textbf{0.885}	\\
     COIL-20& (1440, 400) & 20 & \textit{0.996} & 0.986 & 0.972 & 0.975 &  \textbf{0.991} \\
Activity & (5744, 561) & 6 & \textit{0.853} & 0.769 & 0.829 & 0.779 &  \textbf{0.849}	  \\
\bottomrule
\end{tabular}
\vspace{1cm}
\caption{
\textbf{Classification accuracies of feature selection methods using decoder networks as the learner.} Here we show the classification accuracies of the various feature selection methods on six publicly available datasets. Here Fisher refers to the Fisher score, PFA refers to principal feature analysis and \textit{All-Feature} refers to the learner that uses all input features. For each method, we select $k=50$ features and use a 1-hidden layer neural network for classification. All reported values are on a hold-out test set. (Higher is better.)}
\label{table:experiments}
\end{table}

\section{Extension to Unsupervised Feature Selection}
\label{sec:unsupervised}

\subsection{Background}

\begin{figure}
\centering
\begin{subfigure}{}
  \centering
  \includegraphics[width=.3\linewidth]{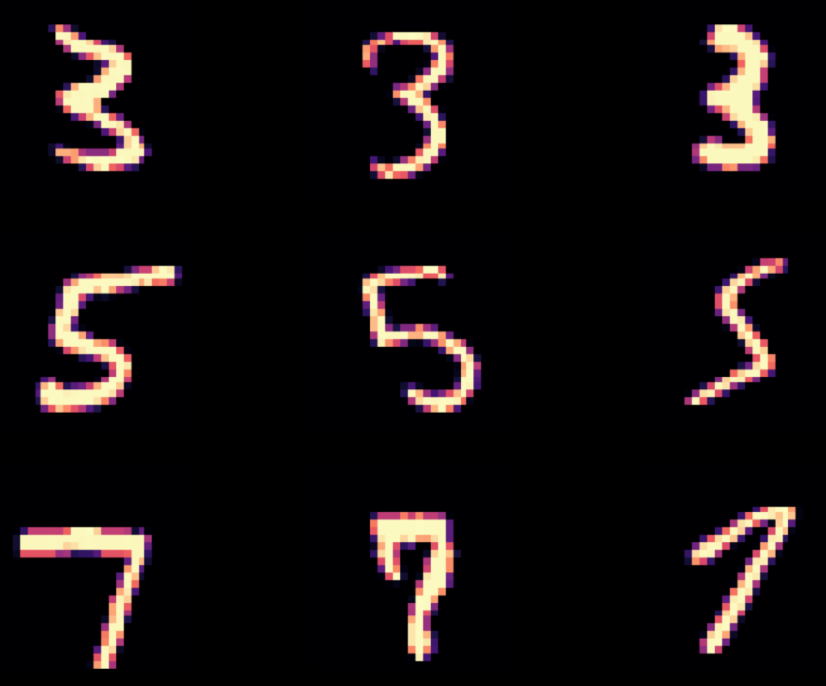}
\end{subfigure}%
\hspace{0.06\linewidth}
\begin{subfigure}{}
  \centering
  \includegraphics[width=.3\linewidth]{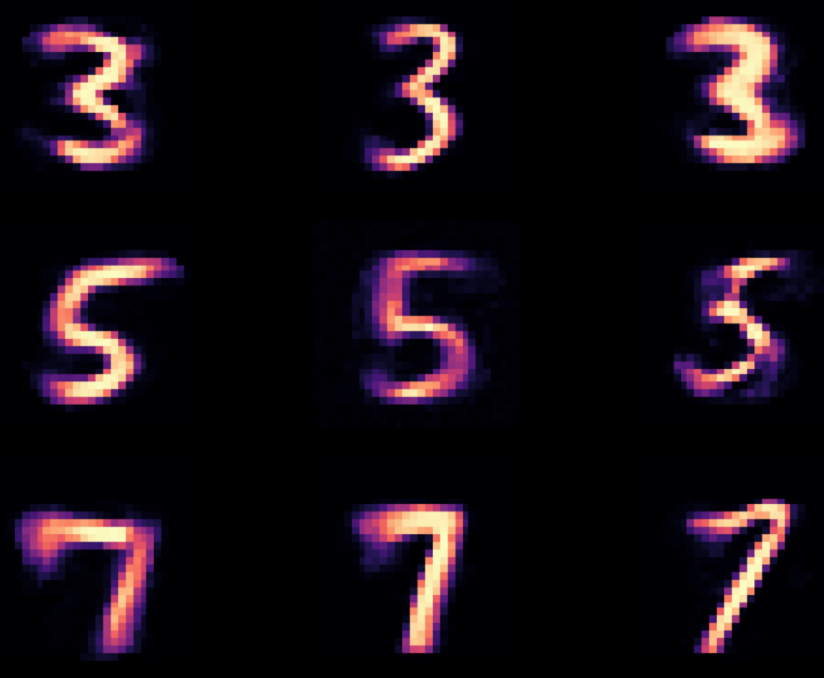}
\end{subfigure}
\caption{
\textbf{Demonstrating the unsupervised LassoNet on the MNIST dataset.} \\
\textit{Left:} 3 test images from each class of digits are shown, sampled at random. \\
\textit{Right:} the reconstructed versions of the test images using LassoNet with an intermediate penalty level (corresponding to about 50 active features) show that generally the digit is identified correctly and some stylistic features, such as the orientation in the digit "5" and the thickness in the digit "7", are preserved. (see Figures \ref{fig:mnist-unsupervised1},\ref{fig:mnist-unsupervised2},\ref{fig:mnist-unsupervised3} which show the results of applying LassoNet to individual classes of digits.)}
\label{fig:reconstructions}
\end{figure}

In certain applications, specific prediction tasks may not be known ahead of time, and thus it is important to develop methods that can identify a subset of features while allowing imputation of the remaining features with minimal distortion for arbitrary downstream tasks. Thus, an \textit{unsupervised} approach becomes relevant in order to identify the most important features in the dataset, and whether there are redundant features that do not need to be measured.

In this section, we show how to adapt LassoNet to the unsupervised setting.
As another potential use case, we note that the algorithm is appropriate for multi-response classification and regression as well.
By selecting a common set of features, the method is able to borrow strength across outputs to impute several different but potentially related responses.

\subsection{Training}

LassoNet adapts to the unsupervised setting easily by replacing the neural network classifier with a decoder network. More precisely, we consider the reconstruction loss $L(\theta,W) = \norm{ f_{\theta, W}(X) - X}_F^2$, where $\norm{\cdot}_F$ denotes the Frobenius matrix norm.

The pseudocode, shown below, is quite similar to that for training the standard LassoNet.
The major difference is the use of the group-LASSO penalty rather than LASSO in order to enforce the same set of selected features across all reconstructed inputs, leading to the \textsc{Group-Hier-Prox} algorithm.

\begin{algorithm}
    \caption{Training LassoNet for Unsupervised Feature Selection}
    \label{alg:unsupervised-lassonet}
	\begin{algorithmic}[1]
	\State \textbf{Input:} training dataset $X \in \mathbb{R}^{n \times d}$ feed-forward neural network $g_W(\cdot)$, number of epochs $B$,  multiplier $M$, path multiplier $\epsilon$, learning rate $\alpha$.
	\State Initialize and train the feed-forward network on the reconstruction loss $L(\theta,W)$
	\State Initialize the penalty, $\lambda=\lambda_0$, and the number of active features, $k=d$
	\While {$k > 0$}
	    \State Update $\lambda \leftarrow (1+\epsilon)\lambda$
	    \For{$b \in \{1 \ldots B\}$}
    	    \State Compute gradient of the loss w.r.t to $\theta$ and $W$ using backpropagation
    	    \State Update $\theta \leftarrow \theta - \alpha \nabla_{\theta}L$ and $W \leftarrow W - \alpha \nabla_{_W}L$
    	    \State Update
    	    $(\theta, W^{(1)}) = 
    	    \textsc{Group-Hier-Prox}
    	    (\theta, W^{(1)}, \alpha \lambda, M)$
	    \EndFor
    	\State Update $k$ to be the number of non-zero coordinates of $\theta$
	\EndWhile
	\end{algorithmic} 
\end{algorithm}

\begin{algorithm}[!ht]
    \caption{Group Hierarchical Proximal Operator}
    \label{alg:group-hier-prox}
	\begin{algorithmic}[1]
	\Procedure{Group-Hier-Prox}{$\theta, W^{(1)};\lambda,M$}
	\State \textbf{Notation:}
	\For{$j \in \{1,\ldots, d\}$}
	    \State Sort the coordinates of $W^{(1)}_{j} \in \R^{K}$ into 
	        $|W^{(1)}_{(j,1)}| \geq \ldots \geq |W^{(1)}_{(j,K)}|$
	    \For{$m \in \{0, \ldots, K\}$}
	        \State
	        $w_{j, m} \gets \frac{M}{1+mM^2} \cdot
	           \st_{\lambda}(\norm{\theta_j}_2 + M \sum_{j=1}^m |W^{(1)}_{j}|)$.
    	\EndFor
    \State Find $\tilde{m}$, the first $m \in \{0,...,K \}$ such that 
    	    $|W^{(1)}_{(j,m)}|\ge w_{j, m} \ge |W^{(1)}_{(j,m+1)}|$.
	\State Update $\Tilde{\theta}_j \gets \frac{1}{M}\cdot w_{j, \tilde{m}} \cdot \frac{\theta_j}{\norm{\theta_j}_2}$
	\State Update $\Tilde{W}^{(1)}_j \gets \sign (W^{(1)}_j)\cdot 
	    \min(w_{j, \tilde{m}},|W^{(1)}_j|)$
	\EndFor
	\State
	\Return $(\Tilde{\theta},\Tilde{W}^{(1)})$
	\EndProcedure
	\State Notation: $d$ denotes the number of features and $K$ the size of the first hidden layer. The vectors $\theta_j \in \R^K$, $W_j^{(1)} \in \R^K$ denote the linear coefficients and the bottom-layer coefficients in the neural networks 
	that are associated with $j$-th feature. 
	\State Conventions:
	Ln. 6, $W^{(1)}_{(j,K+1)} = 0$, $W^{(1)}_{(j,0)} =+\infty$;
	Ln. 10, minimum is applied entry-wise.
    \end{algorithmic}
\end{algorithm}

\subsection{Selected Digits for Single Classes in MNIST}
\begin{figure}[H]
\begin{minipage}{\linewidth}
  \centering
  \includegraphics[width=0.7\linewidth]{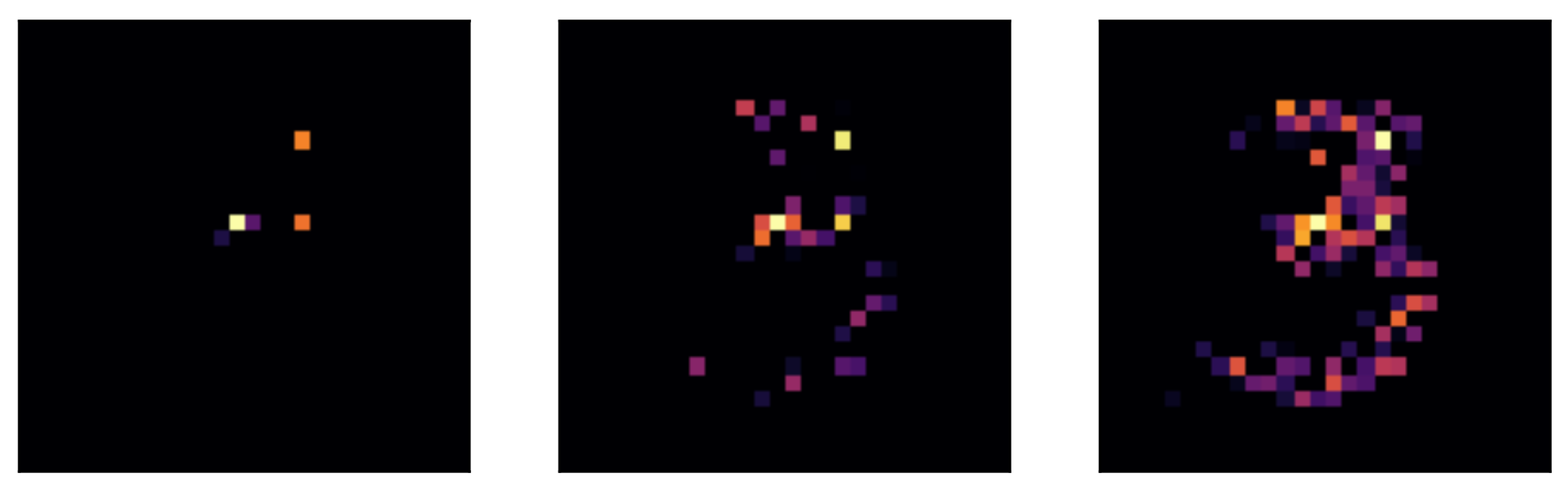}
  \captionsetup{width=.7\linewidth}
  \caption{Results for LassoNet in choosing the most informative pixels of images of the digit 3 in the MNIST dataset, for three different penalty levels ($\lambda = 5,\lambda = 1, \lambda = 0.1).$}
  \label{fig:mnist-unsupervised1}
  \vspace{.5cm}
\end{minipage}
\end{figure}

\begin{figure}[H]
  \centering
  \includegraphics[width=0.7\linewidth]{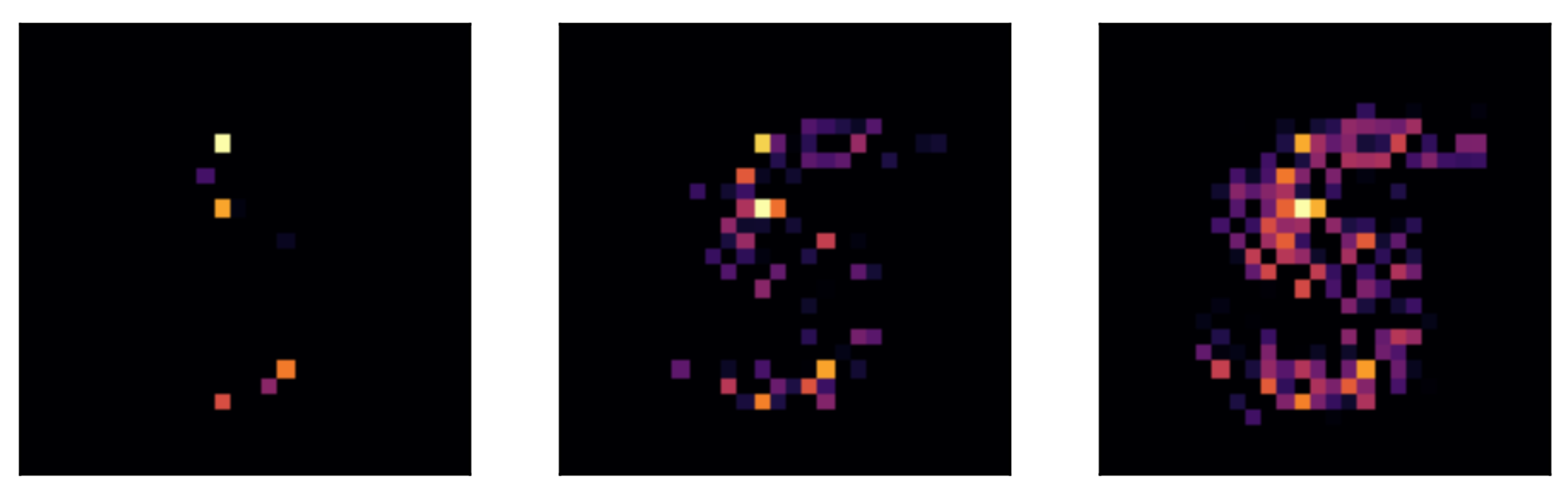}
  \captionsetup{width=.7\linewidth}
  \caption{Results for LassoNet in choosing the most informative pixels of images of the digit 5 in the MNIST dataset, for the three  penalty levels.}
  \label{fig:mnist-unsupervised2}
\end{figure}

\begin{figure}[H]
  \centering
  \includegraphics[width=0.7\linewidth]{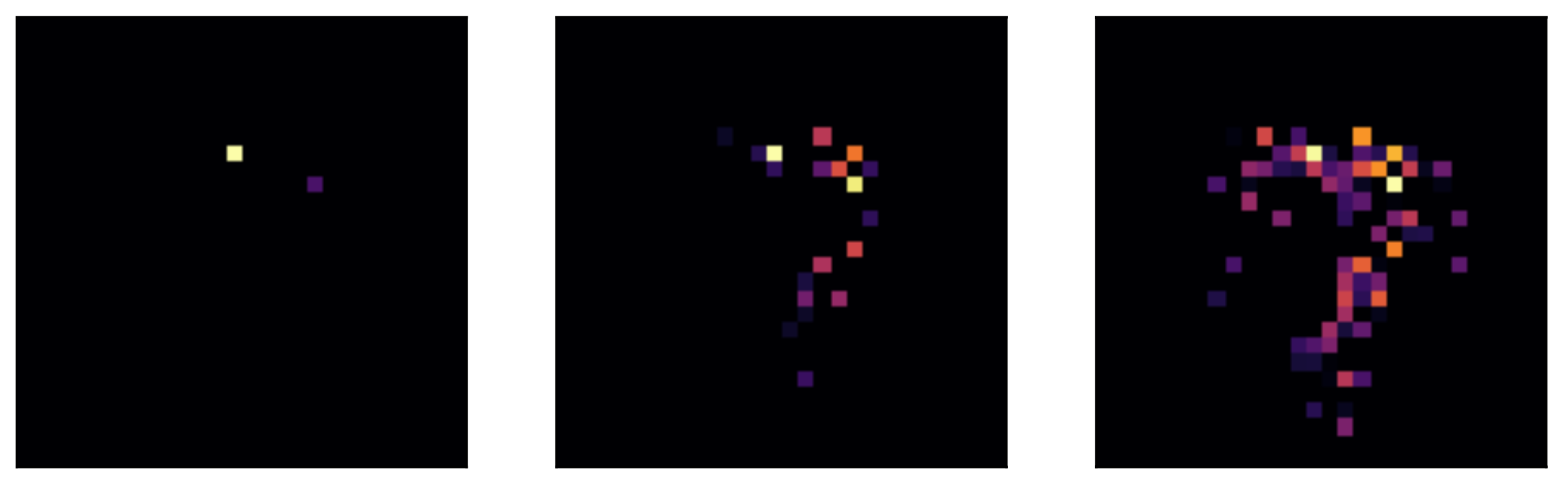}
  \captionsetup{width=.7\linewidth}
  \caption{Results for LassoNet in choosing the most informative pixels of images of the digit 7 in the MNIST dataset, for the three  penalty levels.}
  \label{fig:mnist-unsupervised3}
\end{figure}

We trained LassoNet in this unsupervised manner on subsets of the MNIST data consisting of a single digit. Some representative images for different digit classes are shown in Figs. \ref{fig:mnist-unsupervised1}, \ref{fig:mnist-unsupervised2} and \ref{fig:mnist-unsupervised3}.

\section{ Extension to Matrix Completion}
\label{sec:matrix-completion}

In several problems of contemporary interest, arising for instance in biomedical settings where measurements are costly or otherwise limited, the observed data are in the form of a large sparse matrix, $Z_{ij}, (i, j) \in \Omega$, where  $\Omega \subset \{1,..., m\}\times \{1,..., n\}$.
Popularly dubbed the matrix completion problem 
\citep{candes:recht,MHT2010}, the task is to predict the unobserved entries.

Many existing approaches to the problem \citep{rennie2005fast,bennett2007kdd,srebro2004learning}, including the popular Soft-Impute algorithm \citep{MHT2010} make low-rank assumptions about the underlying matrix.
In this section, we propose an extension of LassoNet that provides a new matrix completion method that does not make any low-rank assumption.
Our method is detailed in Algorithm \ref{alg:lassonet-impute}, and differs from Soft-Impute in two major aspects: 
\begin{itemize}
    \item First, it trains a feed-forward neural network on the imputation task.
    Our method is also iterative and updates the input using the reconstruction from the feed-forward network.
    Here the reconstruction loss is $\norm{X-Z}_F^2$, where $\norm{\cdot}_2$ denotes the Frobenius norm.
    In Soft-Impute, the corresponding operation is singular value thresholding (that is, singular value decomposition followed by soft-thresholding) and finds a linear low-dimensional structure. 
    In contrast, our method uses a nonlinear low-dimensional manifold through the network's hidden layers.
    \item Once the dense model has been trained, the method performs feature selection using the \textsc{Group-Hier-Prox} operator (presented in Algorithm \ref{alg:group-hier-prox}). This allows to prune the original imputation model so that it only uses a small set of input features.
\end{itemize}

\newcommand{\pred}{{\rm pred}}
\newcommand{\old}{{\rm old}}
\begin{algorithm}
    \caption{LassoNet for Matrix Completion}
    \label{alg:lassonet-impute}
	\begin{algorithmic}[1]
	\Procedure{}{}
	\State Input: data matrix $Z$, index of observed entries $\Omega$.
	\State Perform initial imputation
	    \begin{equation*}
	        X[i, j] = \begin{cases}
	                    Z[i, j]~~&\text{if $(i, j) \in \Omega$} \\
	                    \frac{1}{|j: (i, j) \in \Omega|} \sum_{j:(i, j) \in \Omega} Z[i, j]~~&\text{if $(i, j) \not\in \Omega$} 
	              \text{  [row-mean imputation]}
	                  \end{cases}
	    \end{equation*} 
	\While{$X$ not converged}
	    \State Compute $X_{proj} = P_{\Omega}(Z) + P_{\Omega^{\perp}}(X)$
	    \State Train the feed-forward network on the reconstruction loss $L(\theta,W)$ between $X_{proj}$ and $Z$
	    \State Update $X \leftarrow f_{\theta, W}(X_{proj}) $
	\EndWhile
	\State Perform feature selection on the $\lambda$-path using Algorithms \ref{alg:lassonet} [Lns. 4-12] and \ref{alg:group-hier-prox}.
	\EndProcedure

	\State Notation: Ln. 5: $P_{\Omega}(X)$ denotes the projection of $X$ onto the entries in $\Omega$: 
	    \begin{equation*}
	        P_{\Omega}(X)[i, j] = 
	        \begin{cases} 
	           X[i, j]~~&\text{if $(i, j) \in \Omega$} \\
	           0~~&\text{if $(i, j) \not\in \Omega$}
	       \end{cases}
	    \end{equation*}
    \end{algorithmic}
\end{algorithm}

\begin{figure}
\begin{minipage}{.5\textwidth}
  \centering
  \includegraphics[width=1\linewidth]{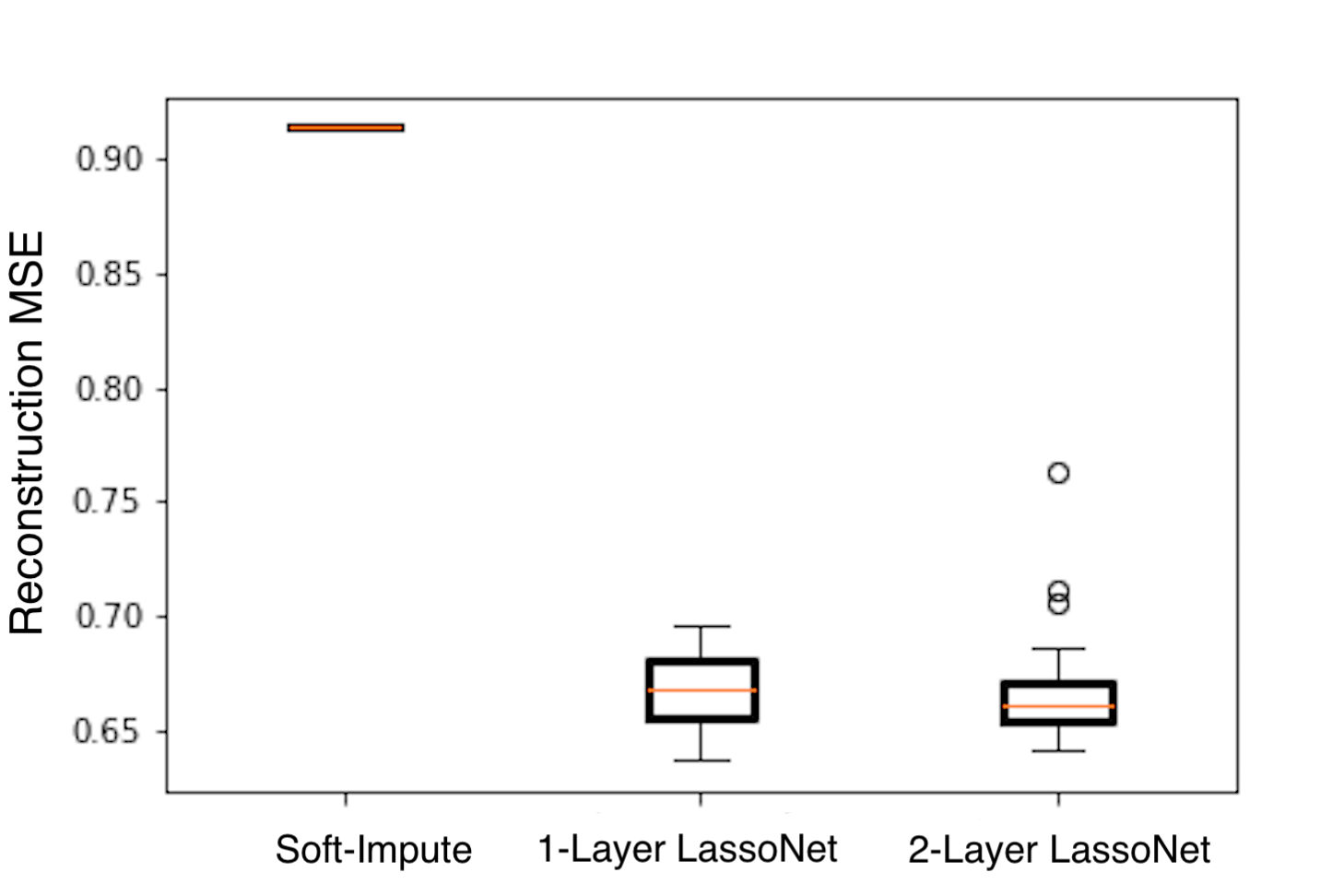}
\end{minipage}\hfill
\begin{minipage}{.5\textwidth}
  \centering
  \includegraphics[width=1\linewidth]{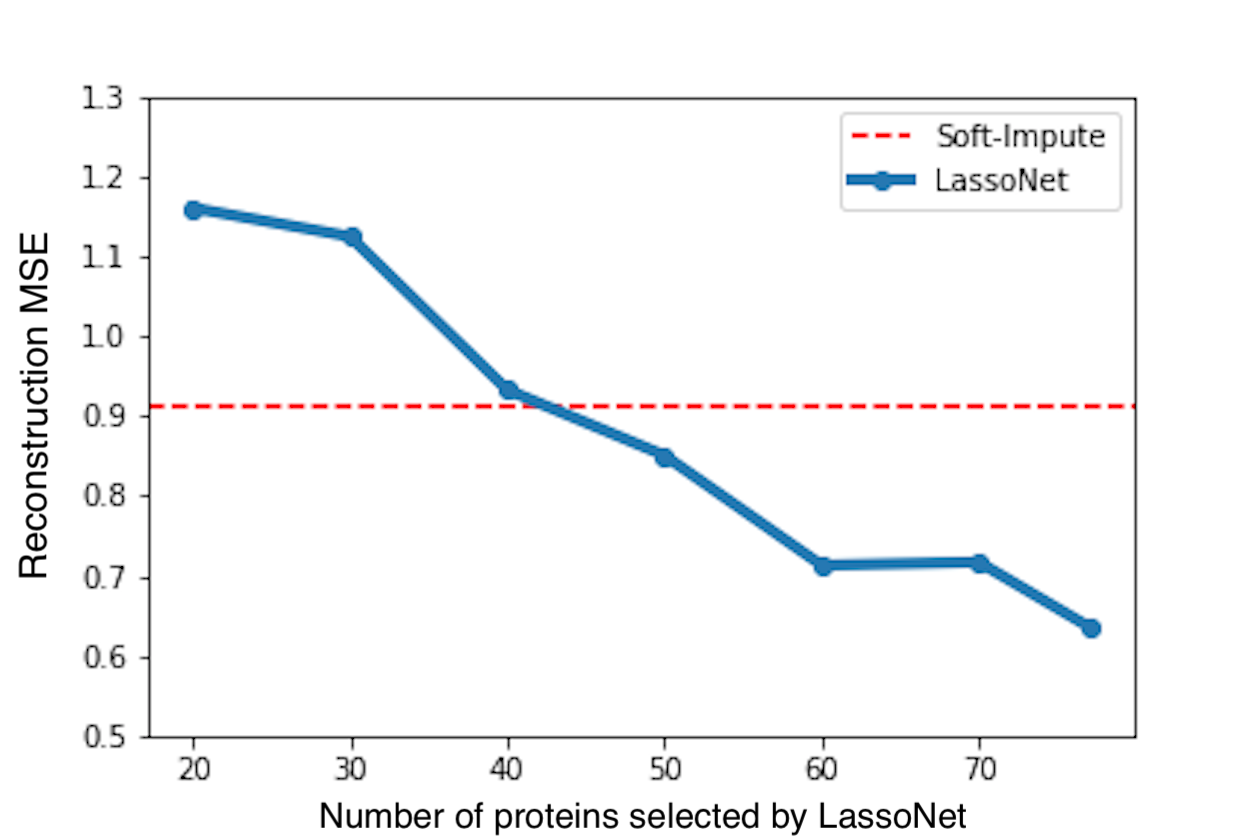}
\end{minipage}%
\caption{
\textbf{Imputation errors of LassoNet and and Soft-Impute.}
Here, we show the mean-squared error of the imputation task on the MICE Protein data set.
\textit{Left:} We report the test MSE for LassoNet and Soft-Impute on the data set prior to performing feature selection.
We observe about a 30\% reduction (note that the y-axis begins at 0.65) of the reconstruction error when using LassoNet. 
Standard deviation bars are shown over 25 trials with different initializations. 
\textit{Right:} We report the performance of LassoNet with different numbers of selected features using the MSE on the test set. 
We find that we can achieve a similar MSE to Soft-Impute using only about 40 proteins, a 50\% reduction in the number of proteins measured.
}
\label{fig:completion}
\end{figure}

To investigate the method's performance, we run both LassoNet and Soft-Impute on the MICE Protein Dataset. 
This dataset was previously used in Section \ref{sec:experiments} for the supervised prediction task.
Here the goal is to impute the entries that are missing from the training data.
Initially, the data is complete with no missing entries (up to the few missing values we mentioned in Section \ref{subsec:data}.
We construct our training set by keeping $80\%$ of the entries at random.
We
use $10\%$ of the data for validating hyper-parameters and for determining the stopping criterion. 
Finally, we reserve $10\%$ of the data for the hold-out test test.
To control for the performance of the reconstruction network, we trained each reconstruction network using early stopping with a patience of $10$ epochs.
Similarly to Section \ref{sec:experiments}, we did not use the network that was learned during training, but retrained the reconstruction network from scratch.
The results are displayed in Figure \ref{fig:completion}, where LassoNet achieves about a 50\% reduction in the number of proteins measured for an equivalent test MSE.
More details on the experimental settings are provided in Appendix \ref{appendix:experiments}.

Our results show that the low-rank assumption underlying most existing matrix imputation methods is not always the most appropriate. 
More worryingly, when the linear assumption is violated, the statistical performance of standard imputation methods may be severely impaired.
Therefore, it may be more reasonable to expect an arbitrary nonlinear low-dimensional structure.
 If that is the case, LassoNet will outperform linear reconstruction.
 
Finally, our method provides the added benefit of feature selection, which is of independent interest when measuring different features is costly or otherwise unavailable.

\section{Sparsity in Learned Features}
\label{sec:other-extensions}

In some problems, the raw features are not interpretable and hence inducing sparsity in these features is not helpful:
\begin{itemize}
    \item For example in computer vision, the inputs are pixels and without perfect registration, a given pixel does not even  have the same meaning across different images.
    In this case, one may instead want to select visual shapes or patterns. The current practice of convolutional neural networks shies away from hand-engineered features and learns these from the data instead \citep{behnam}. 
    Currently, the state-of-the-art is based on learning these convolutional filters. 
    In this setting, it would be desirable to achieve ``filter sparsity”, that is, select the most relevant convolutional filters.
    This can improve the interpretability of the model and alleviate the need for architecture search. 
    In future work, we plan to achieve this by applying LassoNet to the output of one or more  convolutional layers.

    \item Another application area is that of auto-encoders, whose bottleneck layer learns the most representative latent variables. Again, it is often of interest to further reduce the complexity of the auto-encoder by selecting the most important hidden units.
    We believe this is another promising direction for future research. 
\end{itemize}

\section{Discussion}
\label{sec:discussion}
In this paper, we have proposed a new feature selection method for neural networks.
Unlike most other feature selection methods, our method is data-driven and provides a path of regularized models at a cost that is essentially that of training a single model.
At its core, LassoNet involves a nonconvex optimization problem with hierarchy constraints to satisfy feature sparsity. By using proximal gradient descent, the nonconvex optimization problem is decomposed into two subproblems that are solved iteratively, one using stochastic gradient descent and the other analytically. The stochasticity of the initial dense model allows it to efficiently explore and converge over an entire regularization path with varying number of input features. This makes LassoNet different from many feature selection methods, which assume prior knowledge of the number of features to select.

Advantages of LassoNet include its generality and ease of use. 
First, the generality of the method allows it to extend to several other learning tasks, such as unsupervised reconstruction and matrix completion. 
Second, implementing the architecture in popular machine learning frameworks requires only modifying a few lines of code from a standard feed-forward neural network.
Furthermore, the runtime of LassoNet over an entire path of feature sizes is similar to that of training a single model and improves with hardware acceleration and parallelization techniques commonplace in deep learning. 
Finally, the only additional hyperparameter of LassoNet is the hierarchy coefficient.
We find that the default value, $M=10$, used in this paper works well for a variety of datasets.

LassoNet, like the other feature selection methods we compared with in this paper, does not provide $p$-values or statistical significance quantification. Features discovered through LassoNet should be validated through hypothesis testing or additional analysis using relevant domain knowledge. In this regard, a growing body of research about hypothesis testing for Lasso \citep{lockhart2014significance,javanmard2014confidence,taylor2015statistical} could serve as a fruitful starting point. 

\medskip

Python code and documentation for LassoNet is available at \url{https://lassonet.ml}, and R code will soon be available in the same website.

\acks{We would like to thank John Duchi and Ryan Tibshirani for helpful comments. Robert Tibshirani was supported by NIH grant 5R01 EB001988-16 and NSF grant 19 DMS1208164.}

\medskip
\bibliography{20-848}
\bibliographystyle{abbrvnat}

\newpage
\appendix
\section{Additional Experiments}
\label{app:experiments}
In this section, we report the results for different numbers of features on the other data sets, similar to what was done for the ISOLET data set in Figure \ref{fig:isolet}.
These figures complement Table \ref{table:experiments} where the number of features was fixed to $k=50$.
Overall, we find that LassoNet continues to achieve high (though not always highest) accuracy.

We note that for all methods other than LassoNet, the computational cost of producing these figures was multiples of that of producing the table. 
This is because most other feature selection methods take as input $k$, the number of features to select, and need to run from scratch for diferent values of $k$.
On the other hand, for LassoNet the cost remains essentially unchanged.
We refer the reader to Section \ref{subsec:complexity} for further discussion.

\begin{figure}[htb]
    \centering
    \includegraphics[width=1.\linewidth]{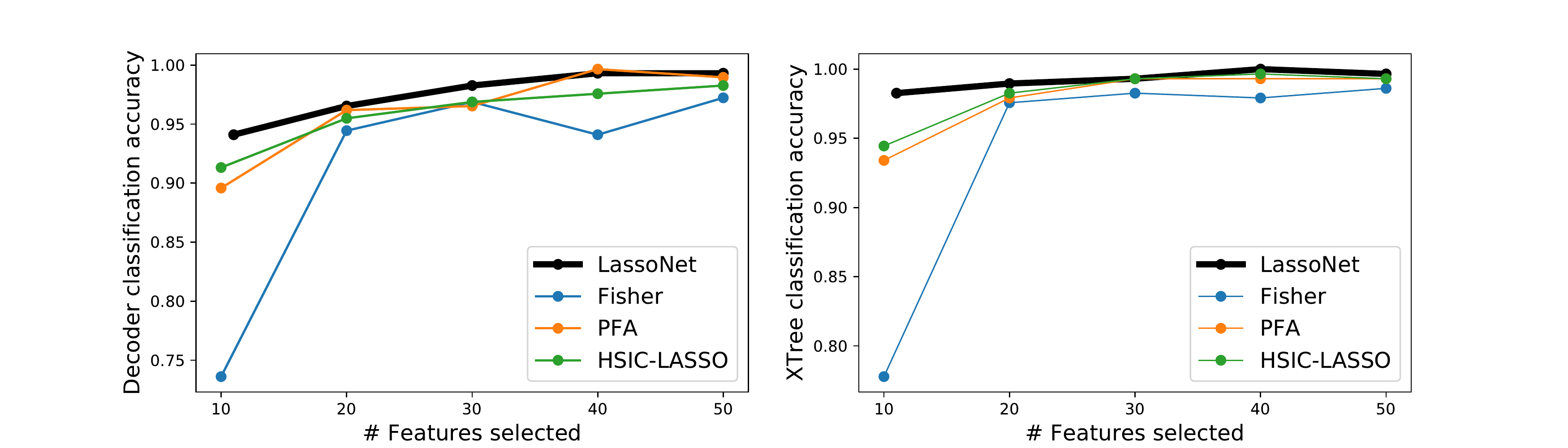}
    \caption{Results on the COIL dataset.}
\label{fig:coil}
\end{figure}

\begin{figure}[htb]
    \centering
    \includegraphics[width=1.\linewidth]{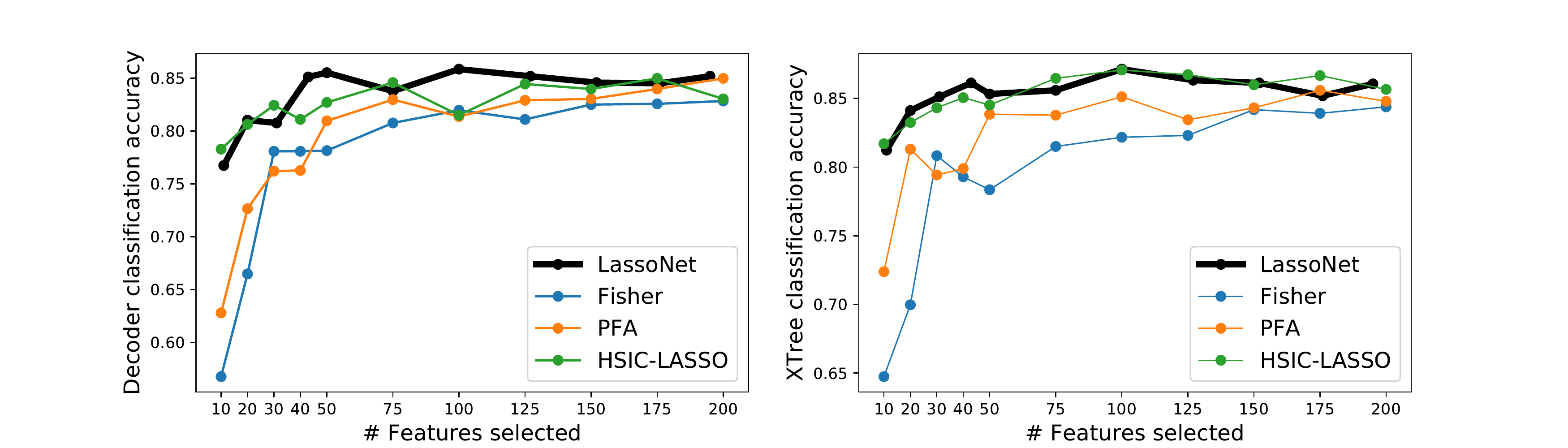}
    \caption{Results on the ACTIVITY dataset.}
\label{fig:activity}
\end{figure}

\begin{figure}[htb]
    \centering
    \includegraphics[width=1.\linewidth]{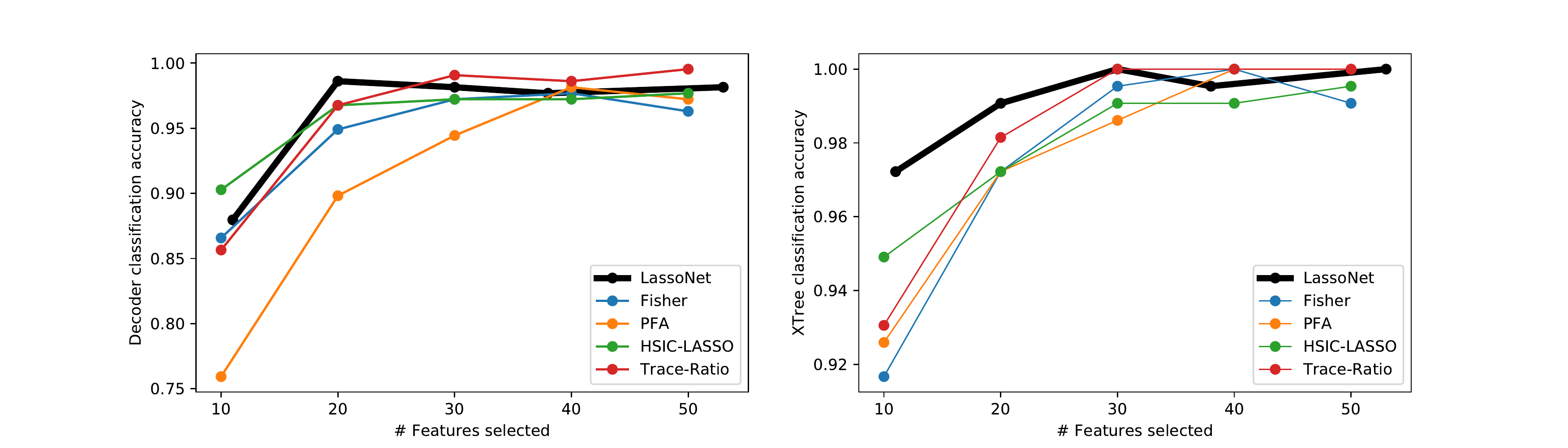}
    \caption{Results on the MICE dataset.}
\label{fig:mice}
\end{figure}

\begin{figure}[htb]
    \centering
    \includegraphics[width=1.\linewidth]{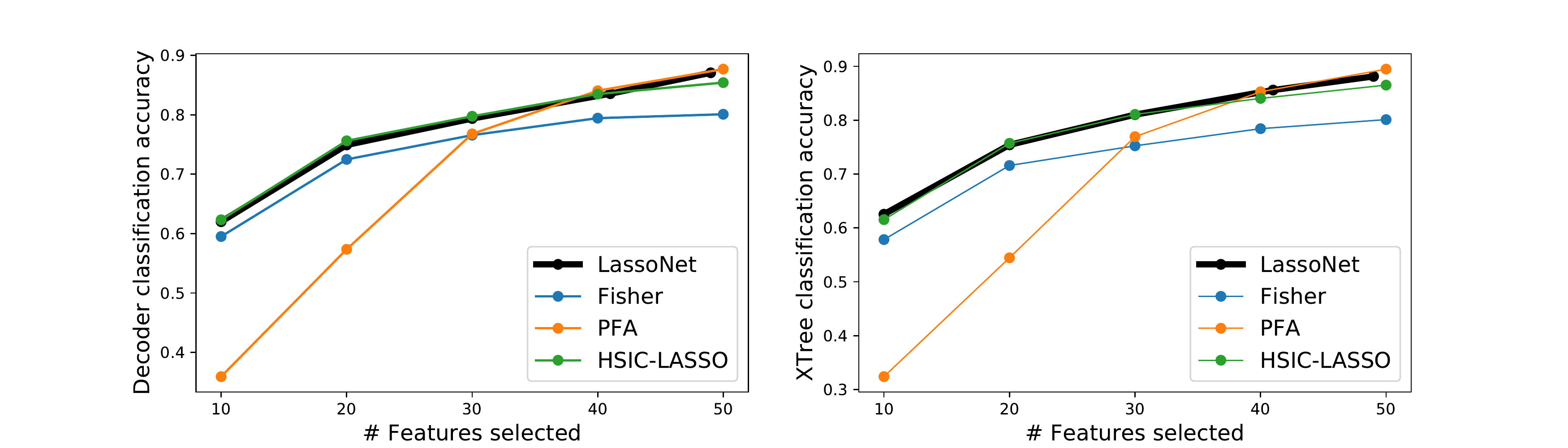}
    \caption{Results on the MNIST dataset.}
\label{fig:mnist-path}
\end{figure}

\section{Proofs}
\subsection*{Proof of Correctness of the \textsc{Hier-Prox} and \textsc{Group-Hier-Prox} Operators}
\label{proof}

\newcommand{\Udata}{U}
\newcommand{\Vdata}{v}
\newcommand{\ub}{b}
\newcommand{\uW}{W}

At its core, LassoNet performs a step of vanilla gradient descent and subsequently solves a constrained minimization problem.
Since the problem is decomposable across features, each iteration of the algorithm decouples into $d$ single-feature optimization problems.
Here we show the following: 
\begin{enumerate}
    \item \textsc{Hier-Prox} returns the global optimum of the following optimization problem:
        \begin{equation}
        \label{eqn:opt-prob-infty}
        \begin{split}
        &\minimize_{b \in \R, W \in \R^K}~~
        	\half (v-b)^2 + \half \ltwo{u-W}^2 + \lambda |b|, \\
        &\subjectto ~\norm{W}_\infty \le  M |b|
        \end{split}
        \end{equation}
        where $v \in \R$ is a scalar and $u \in \R^K$ is a vector. 
    \item \textsc{Hier-Prox-Group} returns the global optimum of the following problem:
        \begin{equation}
            \label{eqn:hier-prox-group-eqn}
        \begin{split}
            &\minimize_{(\ub, \uW)} 
	            ~~\half \left(\ltwo{\Vdata- \ub}^2 + \norm{\Udata - \uW}_{2}^2\right) + \lambda \norm{\ub}_2, \\
            &\subjectto ~\norm{\uW}_\infty \le  M \norm{\ub}_2
        \end{split}
        \end{equation}
        where $\Vdata \in \R^K$, $\Udata \in \R^{K}$ are vectors of the same size.  
\end{enumerate}

It turns out that the aforementioned two results are special cases of the 
following proposition. They can be easily recovered by setting $\bar{\lambda} = 0$.
\begin{proposition}
Fix $\Vdata \in \R^k$ and $\Udata \in \R^{K}$. (Note: the two integers $k, K$ can be different) Let us consider the problem 
\begin{equation*}
\begin{split}
\minimize_{(\ub, \uW)} 
	&~~\half \left(\ltwo{\Vdata- \ub}^2 + \norm{\Udata - \uW}_{2}^2\right) + \lambda \norm{\ub}_2 + \bar{\lambda} \norm{\uW}_1 \\
\subjectto  
	&~~\norm{\uW}_{\infty} \le M \cdot \ltwo{\ub}.
\end{split}
\end{equation*}
We derive the sufficient and necessary condition for characterizing the global optimum $(\ub\opt, \uW\opt)$ of the
above optimization problem: 
\begin{enumerate}
\item Let us order the coordinates 
        $\left\{|\Udata_{i}|\right\}_{i \in [K]}$ in decreasing order 
	\begin{equation*}
		\left|\Udata_{(1)}\right| \ge \left|\Udata_{(2)}\right| \ge \ldots \ge \left| \Udata_{(K)}\right|.
	\end{equation*}
	and define for each $s\in [K] = \{0, 1, \ldots, K\}$ the value $\ub_s$ by 
	\begin{equation*}
	\ub_s = \frac{1}{1+sM^2} \left(1- \frac{a_s}{\ltwo{\Vdata}}\right)_+ \Vdata
		~~\text{where}~~a_s = \lambda -  M\sum_{i=1}^s\left( |\Udata_{(i)}| - \bar{\lambda}\right).
	\end{equation*}
	Then $\ub\opt = \ub_{s\opt}$ where $s\opt \in [K]$ is the unique $s \in [K]$ such that  
	\begin{equation*}
	M\ltwo{\ub_{s}} \in \left[\st_{\bar{\lambda}}(|\Udata_{(s +1)}|), \st_{\bar{\lambda}}(|\Udata_{(s)}|)\right).
	\end{equation*}
	By convention $\st_{\bar{\lambda}}(|\Udata_{(K+1)}|) = 0$ and $\st_{\bar{\lambda}}(|\Udata_{(0)}|) = \infty$. 
\item The $\uW\opt$ must satisfy 
	\begin{equation*}
		\uW \opt = \sign(\Udata) \min\left\{M\ltwo{\ub\opt}, \st_{\bar{\lambda}}(|\Udata|)\right\}.
	\end{equation*}
\end{enumerate}
\end{proposition}

\begin{proof}
We start by proving the claim below: for some $w \in \R_+$
	\begin{equation}
	\label{eqn:claim-infty}
	\text{Claim}:
		\uW \opt = \sign(\Udata) \min\left\{M\ltwo{\ub}, \st_{\bar{\lambda}}(|\Udata|)\right\}.
	\end{equation}
	Denote $w = M\ltwo{\ub\opt}$. By definition, $\uW\opt$ is the minimum of the below optimization problem: 
	\begin{equation*}
	\begin{split}
	&\minimize_{\uW}~~ \half \norm{\Udata - \uW}_F^2 + \bar{\lambda} \norm{\uW}_1 \\
	&\subjectto~~ \norm{\uW}_\infty \le M\ltwo{\ub\opt} = w. 
	\end{split}
	\end{equation*}
	Now that strong duality holds since Slater's condition holds. 
	Thereby, we know for some dual variable $s \in \R_+^{K}$, $\uW\opt$ minimizes the Lagrangian 
	function below: 
	\begin{align*}
	\begin{split}
	\uW\opt &~=\argmin_{\uW \in \R^{K}}  \half \norms{\Udata-\uW}_2^2 +  
		    \sum_{j=1}^K s_{j} |\uW_{j}|
			+ \bar{\lambda} \norm{\uW}_1.
	\end{split}
	\end{align*}
	Now, let's take subgradient and get that $\uW\opt$ needs to satisfy, 
	\begin{equation}
	\label{eqn:KKT-condition-sub}
	\begin{split}
		&\uW\opt_{j} - \Udata_{j} + (\bar{\lambda} + s_{j}) v_{j}\opt = 0~~\text{for some $v_{j}\opt \in \partial (|\uW_{j}\opt|)$}. \\
		&s_{j} (|\uW_{j}\opt| - w) = 0. \\
		&s_{j} \ge 0, ~~\text{and}~~|\uW_{j}\opt| \le w. 
	\end{split}
	\end{equation}
	Now we divide our discussion into two cases: 
	\begin{enumerate}
	\item $s_{j} = 0$. The KKT condition~(Eq.~\eqref{eqn:KKT-condition-sub}) shows 
	$\Udata_{j} = \uW\opt_{j} + \bar{\lambda} v_{j}\opt$ for some $v_{j}\opt \in \partial (|\uW_{j}\opt|)$. This implies that $\uW_{j}\opt = \st_{\bar{\lambda}}(\Udata_{j})$, which is possible if and only if 
	$\left|\st_{\bar{\lambda}}(\Udata_{j})\right| \le w$.
	\item $s_{j} > 0$. The KKT condition~(Eq.~\eqref{eqn:KKT-condition-sub}) gives 
	$|\uW_{j}\opt| = w$. Since 
	$\Udata_{j} = \uW_{j}\opt + (\bar{\lambda} + s_{j}) v_{j}\opt$ for 
	$v_{j}\opt \in \partial (|\uW_{j}\opt|)$, it implies
	$\sign(v_{j}\opt) = \sign(\uW_{j}\opt)= \sign(\Udata_{j})$. 
	Hence $\uW_{j}\opt = \sign(\Udata_{j}) w$. Note if $w \neq 0$, then we must have $v_{j}\opt = \sign(\uW_{j}\opt) = \sign(\Udata_{j})$. Thus, having some $s_{j} > 0$ with 
	$\Udata_{j} = \uW_{j}\opt + (\bar{\lambda} + s_{j}) v_{j}\opt$ is equivalent to that 
	$|\st_{\bar{\lambda}}(\Udata_{i, j})| > w$. 
	\end{enumerate}
	Summarizing the above discussion, we see that $\uW\opt$ must satisfy 
	\begin{equation*}
		\uW_{j}\opt = \sign(\Udata_{j}) \min\{w, \st_{\bar{\lambda}}(|\Udata_{j}|)\}.
	\end{equation*}
	This proves the claim at Eq~\eqref{eqn:claim-infty}. Introduce the mapping 
	$\uW: \R^k \to \R^{K}$
	\begin{equation*}
		\uW_{j}(\ub) = \sign(\Udata_{j}) \cdot \min\left\{M \ltwo{\ub}, \st_{\bar{\lambda}}(|\Udata_{j}|)\right\}.
	\end{equation*}
	The claim at Eq~\eqref{eqn:claim-infty} shows that it suffices to find $\ub$ that minimizes 
	\begin{equation*}
		F(\ub) = 
			\half \left(\ltwo{\Vdata - \ub}^2 + \norm{\Udata - \uW(\ub)}_F^2\right) + 
				\lambda \ltwo{\ub} + \bar{\lambda} \norm{\uW(\ub)}_1. 
	\end{equation*}
	Denote $w = M\ltwo{\ub}$. Order the coordinates $\left\{|\Udata_{i}|\right\}_{i\in [K]}$ in decreasing order 
	\begin{equation*}
		\left|\Udata_{(1)}\right| \ge \left|\Udata_{(2)}\right| \ge \ldots \ge \left| \Udata_{(K)}\right|.
	\end{equation*}
	Define by convention that $\Udata_{(0)} = \infty$ and $\Udata_{(K)+1} = 0$. 
 	Now, we compute the value $F(\ub)$ when $w = M\ltwo{\ub} 
	\in \left[\st_{\bar{\lambda}}(|\Udata_{(s+1)}|), \st_{\bar{\lambda}}(|\Udata_{(s)}|)\right)$. Indeed, we have 
	\begin{equation*}
	\begin{split}
		F(\ub)
			&= \half (1+sM^2) \ltwo{\ub - \frac{1}{1+sM^2} \Vdata}^2 + \left(\lambda -  
				M\sum_{i=1}^s\left( |\Udata_{(i)}| - \bar{\lambda}\right) \right) \ltwo{\ub} + r_s.
	\end{split}
	\end{equation*}
	where $r_s$ denotes the remainder term, which is independent of $\ub$ but can be dependent of $\Udata$, 
	$\Vdata$, $M$, $s$, $\lambda$, $\bar{\lambda}$. Now, let's define for $s\in [K]$
	\begin{equation*}
	F_s(\ub) =  \half (1+sM^2) \ltwo{\ub - \frac{1}{1+sM^2} \Vdata}^2 + \left(\lambda -  
				M\sum_{i=1}^s\left( |\Udata_{(i)}| - \bar{\lambda}\right) \right) \ltwo{\ub}.
	\end{equation*}
	and denote $\ub_s \in \R^k$ to be the global minimum of $F_s$ on $\R^k$, i.e., 
	\begin{equation*}
	\ub_s = \frac{1}{1+sM^2} \left(1- \frac{a_s}{\ltwo{\Vdata}}\right)_+ \Vdata
		~~\text{where}~~a_s = \lambda -  M\sum_{i=1}^s\left( |\Udata_{(i)}| - \bar{\lambda}\right).
	\end{equation*}
	Now we show the two claims below, which implies the desired proposition. 
	\begin{enumerate}[(a)]
	\item There exists one unique $s\opt \in [K]$ such that 
	\begin{equation}
	\label{eqn:def-sopt}
	M\ltwo{\ub_{s\opt}} \in \left[\st_{\bar{\lambda}}(|\Udata_{(s\opt +1)}|), \st_{\bar{\lambda}}(|\Udata_{(s\opt)}|)\right).
	\end{equation}
	\item The global minimum $\ub\opt = \ub_{s\opt}$.
	\end{enumerate}
	\paragraph{Proof of Point $(a)$} Denote $s_{\max} \in [K]$ to be the one that satisfies
	\begin{equation*}
	\st_{\bar{\lambda}}(|\Udata_{(s_{\max}+1)}|) = 0 < \st_{\bar{\lambda}}(|\Udata_{(s_{\max})}|).
	\end{equation*}
	It suffices to prove the existence and uniqueness of $s\opt \in [s_{\max}]$ satisfying Eq~\eqref{eqn:def-sopt}.
	Introduce the function $h: [s_{\max}] \to \R$ by 
	\begin{equation*}
		h(s) = M(\ltwo{\Vdata} - \lambda) + \bar{\lambda} -  |\Udata_{(s)}| + M^2 \sum_{i=1}^s (|\Udata_{(i)}| - |\Udata_{(s)}|) .
	\end{equation*}
	It is clear that $h$ is increasing w.r.t. $s \in [s_{\max}]$. Moreover, by algebraic manipulation, one can show 
	that $s \in [s_{\max}]$ satisfies
	\begin{equation*}
		M\ltwo{\ub_{s}} \in \left[\st_{\bar{\lambda}}(|\Udata_{(s +1)}|), \st_{\bar{\lambda}}(|\Udata_{(s)}|)\right).
	\end{equation*}
	if and only if (set by convention $h(0) = -\infty$, $h (s_{\max}+1) = \infty$)
	\begin{equation*}
		h(s) \le 0 < h(s+1). 
	\end{equation*}
	This proves the existence and uniqueness of $s\opt$ that satisfies Eq~\eqref{eqn:def-sopt}. 

	\paragraph{Proof of Point $(b)$} Introduce the function
	\begin{equation*}
		f(w) = \min_{\ub: M\ltwo{\ub} = w} F(\ub).
	\end{equation*}
	We will show that 
	\begin{enumerate}[(i)]
	\item $f(w)$ is strictly increasing when $w \in  \left[\st_{\bar{\lambda}}(|\Udata_{(s +1)}|), \st_{\bar{\lambda}}(|\Udata_{(s)}|)\right)$
		and $s < s\opt$.
	\item $f(w)$ is strictly decreasing when $w \in  \left[\st_{\bar{\lambda}}(|\Udata_{(s +1)}|), \st_{\bar{\lambda}}(|\Udata_{(s)}|)\right)$
		and $s > s\opt$.
	\end{enumerate}
	The above two facts imply that the global minimum $w\opt$ of $f(w)$ must belong to the interval 
	$w \opt \in \left[\st_{\bar{\lambda}}(|\Udata_{(s\opt +1)}|), \st_{\bar{\lambda}}(|\Udata_{(s\opt)}|)\right]$. Thus 
	$\ub = \ub_{s\opt}$ is the global minimum of $F(\ub)$ since 
	$\ub = \ub_{s\opt}$ achieves the minimum over all $\ub$ that satisfy
	$w = M\ltwo{\ub} \in \left[\st_{\bar{\lambda}}(|\Udata_{(s\opt +1)}|), \st_{\bar{\lambda}}(|\Udata_{(s\opt)}|)\right)$.
	
	Now we prove point $(i)$ and point $(ii)$. For $w \in  \left[\st_{\bar{\lambda}}(|\Udata_{(s +1)}|), \st_{\bar{\lambda}}(|\Udata_{(s)}|)\right)$, 
	we know that $F(\ub) = F_s(\ub) + r_s$. From all $\ub$ satisfying $M\ltwo{\ub} = w$, it is clear 
	that $\ub = \frac{w\Vdata}{M\ltwo{\Vdata}}$ minimizes $F_s(\ub)$. Therefore, for 
	$w \in  \left[\st_{\bar{\lambda}}(|\Udata_{(s +1)}|), \st_{\bar{\lambda}}(|\Udata_{(s)}|)\right)$, 
	\begin{equation*}
		f(w) =F\left(\frac{w\Vdata}{M\ltwo{\Vdata}}\right) = \frac{1+sM^2}{2M^2} \left(w - w_s\right)^2 
			+ r_s^\prime~~\text{for $w_s = \frac{M(\ltwo{\Vdata} - a_s)}{1+sM^2}$},
	\end{equation*}
	where $r_s$ denotes the remainder term, which is independent of $\ub$ but can be dependent of $\Udata$, 
	$\Vdata$, $M$, $s$, $\lambda$, $\bar{\lambda}$. Now, by definition of $s\opt$, we know that 
	\begin{enumerate}[(1)]
	\item $h(s) \le 0$ for $s \le s\opt$.
	\item $h(s) > 0$ for $s > s\opt$. 
	\end{enumerate}
	Simple algebraic manipulation shows that, this implies that 
	\begin{enumerate}[(1)]
	\item $w_s \le \st_{\bar{\lambda}}(|\Udata_{(s +1)}|)$ for $s < s\opt$. 
	\item $w_s > \st_{\bar{\lambda}}(|\Udata_{(s )}|) $ for $s > s\opt$. 
	\end{enumerate}
	Note that $f(w)$ is quadratic centered at $w = w_s$ for $w \in  \left[\st_{\bar{\lambda}}(|\Udata_{(s +1)}|), \st_{\bar{\lambda}}(|\Udata_{(s)}|)\right)$.
	This proves the desired deferred point $(i)$ and point $(ii)$.
\end{proof}

\newpage
\section{Classification accuracies using tree-based classifiers as the downstream learner}
\label{table:experiments_tree}
Here we report the equivalent table to Table \ref{table:experiments}, but using the extremely randomized tree classifier as downstream learner.
The results are competitive, as LassoNet continues to have high (but not always highest) classification accuracy.

\vspace{1cm}
\begin{table}[!htbp]
  \centering
  \begin{tabular}{cccccccc}
 \midrule
    Dataset & $(n,d)$ & \# Classes & \textit{All-Feature} & Fisher & HSIC-Lasso & PFA & LassoNet\\
    \midrule
    Mice Protein & (1080, 77) & 8 & \textit{0.997} & 0.996 & 0.996 & 0.997 &  \textbf{0.997} \\
    MNIST & (10000, 784) & 10 & \textit{0.941} &  0.818 & 0.869 & 0.879 &  \textbf{0.892}	 \\	
    MNIST-Fashion & (10000, 784) & 10 & \textit{0.831} &  0.66 & 0.775 & 0.784 & \textbf{0.794} \\	
     ISOLET & (7797, 617) & 26 & \textit{0.951} & 0.818 & 0.888 & 0.855&  \textbf{0.891}	  \\
     COIL-20& (1440, 400) & 20 & \textit{0.996} &  \textbf{0.996} & 0.993 & 0.993 & 0.993 \\
Activity & (5744, 561) & 6 & \textit{0.859} &  0.794 & 0.845 & 0.808 &  \textbf{0.860}	  \\
\bottomrule
\end{tabular}
\vspace{1cm}
\caption{
\textbf{Classification accuracies of feature selection methods using the tree-based learner.} Here, we show the classification accuracies of the various feature selection methods on six publicly available datasets. Here Fisher refers to the Fisher score, PFA refers to principal feature analysis, and \textit{All-Feature} refers to the learner that uses all input features. For each method, we select $k=50$ features. The classifier used here was an Extremely Randomized Tree classifier (a variant of random forests) with the number of trees being 50. All reported values are on a hold-out test set. (Higher is better.)}
\end{table}

\newpage
\section{Experimental Details}
\label{appendix:experiments}
All experiments were run on a single computer with NVIDIA Tesla K80 and Intel Xeon E5-2640. 

\subsection{LassoNet Architecture}
The implementation was conducted in the PyTorch framework.
For LassoNet, we use a one-hidden-layer feed-forward neural network with ReLU activation function.
We also included a two-hidden-layer network in Section \ref{sec:matrix-completion} for the matrix completion problem.
The number of neurons in the hidden layer was varied within $[d/3,2d/3,d,4d/3]$, where $d$ is the total number of features, and the network with the highest validation accuracy was selected and measured on the test set.
We used a learning rate of $0.001$ and early stopping criterion of $10$.
Although the hierarchy parameter could in principle be selected on a validation set as well, we have found that the default value $M = 10$ works well for a variety of datasets.

\subsection{Benchmark Datasets}
The MNIST and MNIST-Fashion datasets were retrieved using their official source. The remaining datasets were retrieved from the UCI Repository \citep{Dua:2019}.

\end{document}